\documentclass{article}
\usepackage[utf8]{inputenc}

\usepackage{amsmath}
\usepackage{amsfonts}
\usepackage[capitalise]{cleveref}
\usepackage{booktabs}
\usepackage{array}
\usepackage{mathtools}
\usepackage{algorithmic}
\usepackage{algorithm}

\usepackage[numbers, square]{natbib}
\usepackage{tikz}
\usetikzlibrary{positioning}
\usetikzlibrary{bayesnet}
\usetikzlibrary{fit}
\usetikzlibrary{shapes,arrows}
\usetikzlibrary{shapes.multipart}
\usetikzlibrary{fit}
\usepackage{graphicx}
\usepackage{bm}
\usepackage{bbm}
\usepackage{caption}
\usepackage{subcaption}
\usepackage{sidecap}
\usepackage[export]{adjustbox}
\usepackage{longtable}
\usepackage{rotating}
\usepackage{url}
\usepackage{authblk}
\usepackage{parskip}

\def\*#1{\mathbf{#1}}

\newcommand{\E}{\mathbb{E}}
\newcommand{\KL}{D_{\mathrm{KL}}}
\newcommand{\pthetax}{p_{\theta_{\*x}}}
\newcommand{\pthetam}{p_{\theta_{\*m}}}
\newcommand{\pthetaz}{p_{\theta_{\*z}}}
\newcommand{\pthetah}{p_{\theta_{\*h}}}
\newcommand{\qphiv}{q_{\oldphi_{\*v}}}
\newcommand{\qphiz}{q_{\oldphi_{\*z}}}
\newcommand{\qphig}{q_{\oldphi_{\*g}}}

\let\oldphi\phi
\renewcommand{\phi}{\ensuremath{\varphi}}

\newcommand{\themethod}[0]{HealthGen}

\title{Conditional Generation of Medical Time Series for Extrapolation to Underrepresented Populations}

\author[1,2,*]{Simon Bing}
\author[3]{Andrea Dittadi}
\author[4,5,6,$\sharp$]{\\Stefan Bauer}
\author[6,$\sharp$]{Patrick Schwab}
\affil[1]{ETH Zürich, Zürich, Switzerland}
\affil[2]{Max Planck Institute for Intelligent Systems, Tübingen, Germany}
\affil[3]{Technical University of Denmark, Copenhagen, Denmark}
\affil[4]{KTH Stockholm, Stockholm, Sweden}
\affil[5]{CIFAR Azrieli Global Scholar, Toronto, Canada}
\affil[6]{GlaxoSmithKline, Artificial Intelligence \& Machine Learning, Zug, Switzerland}
\affil[$\sharp$ ]{Joint senior author}
\affil[*]{Corresponding author: simon.bing95@gmail.com}

\date{}                     %% if you don't need date to appear
\begin{document}

\maketitle

\begin{abstract}
The widespread adoption of electronic health records (EHRs) and subsequent increased availability of longitudinal healthcare data has led to significant advances in our understanding of health and disease with direct and immediate impact on the development of new diagnostics and therapeutic treatment options. However, access to EHRs is often restricted due to their perceived sensitive nature and associated legal concerns, and the cohorts therein typically are those seen at a specific hospital or network of hospitals and therefore not representative of the wider population of patients. Here, we present \themethod{}, a new approach for the conditional generation of synthetic EHRs that maintains an accurate representation of real patient characteristics, temporal information and missingness patterns. We demonstrate experimentally that \themethod{} generates synthetic cohorts that are significantly more faithful to real patient EHRs than the current state-of-the-art, and that augmenting real data sets with conditionally generated cohorts of underrepresented subpopulations of patients can significantly enhance the generalisability of models derived from these data sets to different patient populations. Synthetic conditionally generated EHRs could help increase the accessibility of longitudinal healthcare data sets and improve the generalisability of inferences made from these data sets to underrepresented populations.

\end{abstract}

\section{Introduction}
\label{sec:intro}

The broad use of electronic health records (EHRs) has lead to a significant increase in the availability of longitudinal health care data. As a consequence, our understanding of health and disease has deepened, allowing for the development of diagnostic and therapeutic approaches directly derived from EHR patient data.
Models that utilize rich healthcare time series data derived from clinical practice could enable a variety of use cases in personalised medicine, as evidenced by the numerous recent efforts in this area \citep{trew,heart_rate, covews, kidney}.
However, the development of these novel diagnostic and therapeutic tools is often hampered by the lack of access to actionable patient data \citep{Alaa2021GenerativeTM}. 

Even after being deidentified, EHR data is perceived as highly sensitive and clinical institutions raise legal and privacy concerns over the sharing of patient data they may have access to \cite{data_barriers}. Furthermore, even if data is made public, it often originates from a single institution only \citep{hirid, mimiciii, eicu}, resulting in a data set that may not be representative for more general patient populations. Basing machine learning models on single site data sets only risks overfitting to a cohort of patients that is biased towards the population seen at one clinic or hospital, and renders their use for general applications across heterogeneous patient populations uninformative at best and harmful at worst \citep{ gen_med, pneum}.

Putting aside the issue of non-representative patient cohorts, the development of accurate machine learning-based models for healthcare is further impeded by the imbalance in magnitude of the available data compared to other domains. While fields such as computer vision or language modelling have made significant advances, thanks in part to access to large-scale training data sets like ImageNet \citep{imagenet} or large text corpora derived from the World Wide Web, there do not yet exist any comparable data repositories for machine learning in healthcare that may spur innovation at similar pace. Practical problems may also arise during model development due to a lack of training samples for specific, rare patient conditions. If one wishes to study a model's behaviour given data with certain properties, such as only patients with a certain pre-existing condition, medical data sets may often be too small to representatively cover such populations.

One potentially attractive approach to address the aforementioned issues would be to generate realistic, synthetic training data for machine learning models.  Given access to an underlying distribution that approximates that of the real data, paired with the capability to sample from it, one could theoretically synthesize data sets of any desired size. The generated synthetic patient data can be used for assessing  \cite{ chen2019validity,  chen2021synthetic, tucker2020generating}
or even improving machine learning healthcare based software e.g. for liver lesions classification \cite{frid2018synthetic}. If the generative model of the data were to also have the capacity to generate samples conditioned on factors that may be freely chosen, such as for example pre-existing conditions, data sets with the exact properties required for a specified task could be generated as well. Previous reports suggest that such synthetically generated data sets may furthermore be shared with a significantly lower risk of exposing private information \citep{buczak}.

Developing models with synthetic data is already widely applied in machine learning research. In Reinforcement Learning for example, it is the de-facto standard to train models in simulation, in order to have high-fidelity control over the environment \citep{tremblay2018training,ahmed2021causalworld}, or simply because experiments in the real world would be to costly, unethical or dangerous to conduct. Some previous work even suggests that models trained on synthetic data could outperform those derived from real data sets \citep{dope}. The gap between real and synthetic data is rapidly closing in fields like facial recognition in computer vision, as has for example recently been demonstrated by \citet{wood2021fake}.

Classical approaches to generating medical time series data exist, but they fall short of the requirements that modern data-driven models require for their input. Some works employ hand crafted generation mechanisms followed by expensive post-hoc rectifications by clinical professionals \cite{buczak}, while others rely on mathematical models of biological subsystems such as the cardiovascular system \citep{ecg_ode, reacdiffecg}, which require a detailed physiological understanding of the system to be modelled. When the output data stems from multiple, interconnected subsystems whose global dynamics are too complex to model with ordinary differential equations and the size of the required data set is too large to tediously correct unrealistic samples by experts, these approaches may be difficult to utilize.

A natural approach to learning complex relationships from data is to move away from hand-crafted generative models and utilize machine learning methods. While a plethora of powerful generative models for medical imaging data generation have been brought forward in recent years \citep{overcoming_barriers, dermgan, whole_slide, skandarani2021gans}, relatively little research has been reported on generating synthetic medical time series data \cite{Alaa2021GenerativeTM,  dash2020medical, jarrett2021time, van2021decaf}. Moreover, the generation and evaluation of synthetic patient data \citep{goncalves2020generation} is often challenging due to the high missingness in the original  medical datasets \citep{ma2021identifiable,nabi2020full,rubin1976inference, scheffer2002dealing}.

To address these issues, we present HealthGen, a new approach to conditionally generate EHRs that accurately represent real measured patient characteristics, including time series of clinical observations and missingness patterns. In this work, we demonstrate that the patient cohorts generated by our model are significantly better-aligned to realistic data compared to various state-of-the-art approaches for medical time series generation. We demonstrate that our model outperforms previous approaches due to its explicit development for real clinical time series data, resulting in modelling not only the dynamics of the clinical covariates, but also their patterns of missingness which have been shown to be potentially highly informative in medical settings \citep{grud}.
We show that our model's capability to synthesize specific patient subpopulations by means of conditioning on their demographic descriptors allows us to generate synthetic data sets which exhibit more fair downstream behavior between patient subgroups, than competing approaches.
Moreover, we demonstrate that by conditionally generating patient samples of underrepresented subpopulations and augmenting real data sets to equally represent each patient group, we can significantly boost the fairness\footnote{In this work, we measure fairness in terms of the difference in AUROC score between considered groups.} of downstream models derived from the data. 
Furthermore, we evaluate the quality and usefulness of the data we generate using a downstream task that represents a realistic clinical use-case - allowing us to compare our model against competing approaches in a setting that is relevant for clinical impact.

Our main contributions are:
\begin{itemize}
    \item We introduce HealthGen, a new machine-learning method to conditionally generate realistic EHR data, including patient characteristics, the temporal evolution of clinical observations and their associated missingness patterns over time.
    \item We experimentally show that our method outperforms current state-of-the-art models for medical time series generation in synthetically generating patient cohorts that are faithful to real patient data.
    \item We demonstrate the high fidelity of control over synthetic cohort composition that our model provides by means of its conditional generation capability, by generating more diverse synthetic data sets than competing approaches, which ultimately leads to a more fair representation of different patient populations.
    \item We show that by augmenting real data with conditionally generated samples of underrepresented populations, the models derived from these data sets exhibit significantly more fair behaviour than those derived from  unaltered real data.
    \item We perform a comprehensive computational evaluation in realistic clinical use cases to evaluate the comparative performance of HealthGen against various state-of-the-art generative time-series modelling approaches.
\end{itemize}

\section{Results}
\label{sec:results}

\subsection{Overview}
For this study, we consider the MIMIC-III data set \citep{mimiciii}, which consists of EHRs containing time series of measurements of patients that spent time in the intensive care unit (ICU). Additionally, each patient is described by static variables such as their age, ethnicity, insurance type and sex. The time series of a given patient is labelled to indicate whether or not one of the following clinical interventions was performed: mechanical ventilation, vasopressor administration, colloid bolus administration, crystalloid bolus administration or non-invasive ventilation.

After extracting the cohort of patients from the data base, we split them into training (70\%), validation (15\%) and test (15\%) sets, stratified by their binary intervention labels. The generative models are trained on the real training data \(\mathcal{D}_{\text{train}} = \{\*x^n_{1:T}, \*m^n_{1:T}, \*s^n, \*y^n\}^{N_{\text{train}}}_{n=1}\), where \(\*x_{1:T}\) denotes the time series of covariates, \(\*m_{1:T}\) the time series of binary masks indicating where values are missing, \(\*s\) the static patient variables and \(\*y\) a patient's set of labels for the respective clinical interventions. We include the missingness information \(\*m_{1:T}\) explicitly, as previous work has shown that patterns of missing values are highly informative \citep{rubin1976inference} and especially in the medical setting including them is preferential to imputation \citep{grud}.

To evaluate and compare generative models, we first train a downstream time series classification model developed for medical data \citep{grud} on the data sets synthesized by each model. This classifier that has been trained with synthetic data is evaluated on the held-out real test data, and the resulting AUROC score (Area Under the Receiver Operating Characteristic curve) is compared with the AUROC score of the same classifier trained on the real data. The final measure for how faithful a given generated data set is to the real data is the difference between the evaluation score of the synthetic data and that of the real data. Details on the experimental pipeline can be found in \cref{sec:methods}.

In our first experiment, we generate synthetic patient cohorts that are faithful to the real data in terms of their demographics, i.e. they contain the same number of patients as the original data under the same distribution of static variables. Repeating this synthetic cohort generation for all of the available clinical intervention labels, we compare the performance of our model to baseline models for generating clinical time series data. As baselines, we consider Stochastic Recurrent Neural Networks (SRNN)~\citep{srnn} and KalmanVAE (KVAE)~\citep{kvae}, both based on variational autoencoders (VAE) \citep{vae}, and TimeGAN \citep{timegan}, based on generative adversarial networks (GAN)~\citep{gan}.
We then present results of an extension of the previous experiment demonstrating our proposed approach's conditional generation capability, where we generate patient cohorts with static variable distributions that differ from the real data, and investigate the fairness of models derived from the resulting synthetic data.
In an additional experiment, we identify real data settings where some subpopulations of patients have a significantly lower classification score than other patients. Using the the conditional generation capability of our model, we augment the real data with synthetic samples of minority groups and test if this augmentation leads to an increase in the downstream classification score of the previously underrepresented populations.

\subsection{Generating synthetic patient cohorts}
In the first experiment of this work, we investigate our model's capability to generate synthetic data that is faithful to the real data and useful in downstream tasks. 
To study the generative performance of our model and compare it against competing approaches for medical time series generation, we employ the experimental framework described in detail in the Methods section.

Here, we train each generative model conditioned on the real labels \(\*y\) and then generate a synthetic data set \(\hat{\mathcal{D}}\), where the generation is again conditioned on the label of the considered task, to guarantee that the synthetic data shares the same statistics in terms of split between positive and negative labels as the real data. While we could in practice generate as much data as we wish, we synthesize data sets containing the same number of patients as the real data for this experiment, to facilitate a fair comparison between the real patient cohorts and those that have been synthetically generated.

\begin{sidewaysfigure}
    \centering
    \includegraphics[width=\textwidth]{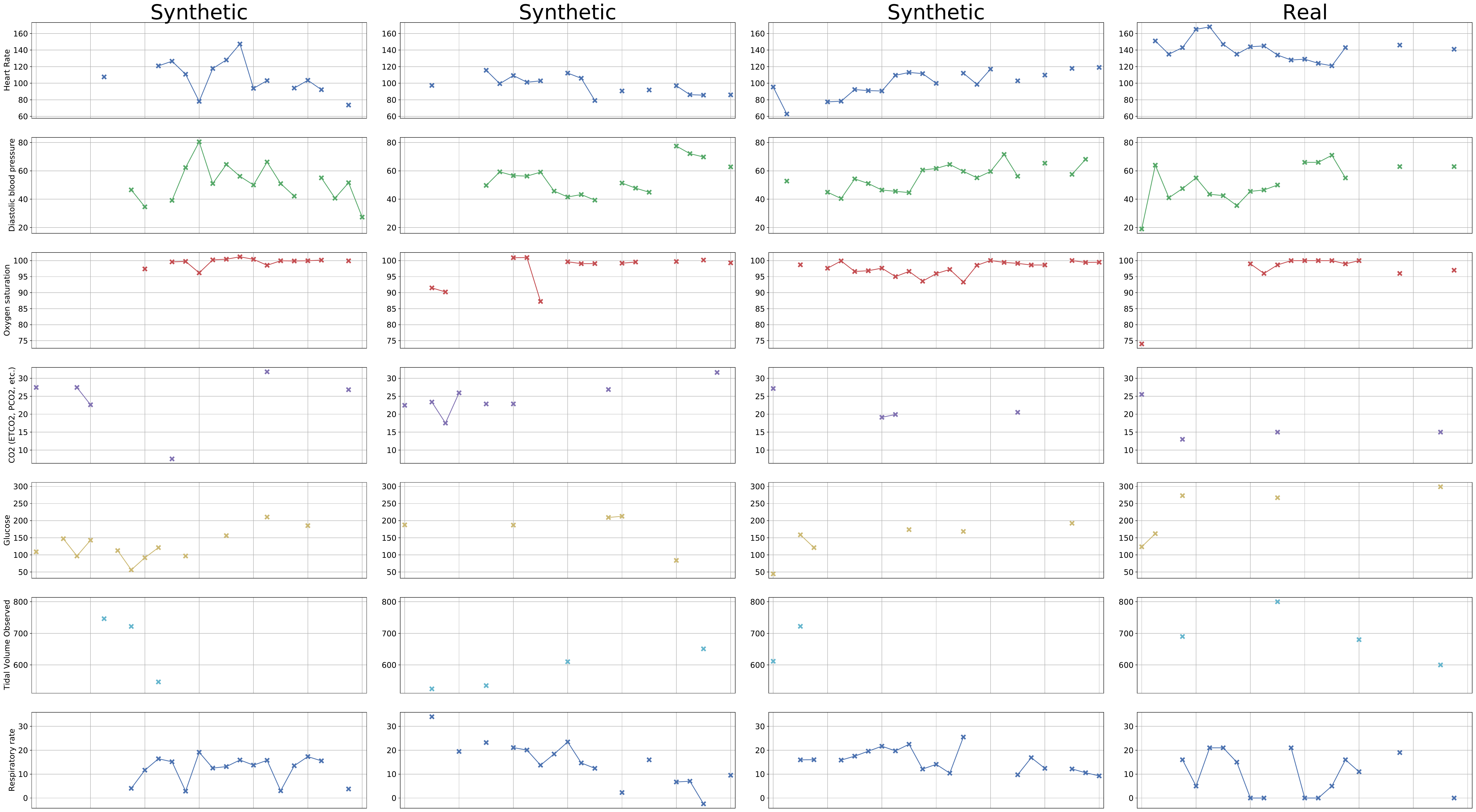}
    \caption{Sample time series of synthetically generated patients, with the time series of one real patient for comparison.}
    \label{fig:mimic_examples}
\end{sidewaysfigure}

A downstream model for medical time series classification \citep{grud} is then trained on the synthetic training data \(\hat{\mathcal{D}}_{\text{train}}\) of each generative model and evaluated on a held out test data set \(\mathcal{D}_{\text{test}}\) consisting of real patients. We compare the performance of our model against the current state-of-the-art models for time series generation, across all five available classification tasks in the MIMIC-III data. The results of these experiments are summarized in \Cref{tab:synth_gen}.
We provide examples of synthetically generated patients' time series in \Cref{fig:mimic_examples}, next to a real patient's time series for comparison. Additional, more detailed visualisations are presented in \Cref{app:synth_examples}.

In all considered tasks, our approach significantly outperforms the state-of-the-art models in synthetically generating medical time series. The fact that our model's downstream classification score is higher then the baselines', across all experimental settings, suggests that the synthetic data generated by our model is more faithful to the real data and thus more useful for the development of downstream clinical time series prediction tasks than the cohorts generated by any of the competing architectures.

\begin{table}[h]
  \centering
   \caption{Comparison of AUROC scores for all predictive tasks between HealthGen and the baseline models. The 95\%  confidence interval of the mean value is presented in parentheses and is estimated via bootstrapping with 30 samples.}
  % \resizebox{<width>}{<height>}{<tabular>}
  \resizebox{\textwidth}{!}{%
    \begin{tabular}{ lccccc }
    \toprule
        & \texttt{vent} & \texttt{vaso} & \texttt{colloid\_bolus} & \texttt{crystalloid\_bolus} & \texttt{niv} \\ 
    \midrule
    Real Data & 0.809 (0.807, 0.811) & 0.801 (0.799, 0.803) & 0.751 (0.741, 0.760) & 0.613 (0.609, 0.616) & 0.634 (0.632, 0.637)  \\
    HealthGen (Ours) & \textbf{0.769 (0.767, 0.772)} & \textbf{0.722 (0.718, 0.727)} & \textbf{0.664 (0.650, 0.678)} & \textbf{0.574 (0.571, 0.577)} & \textbf{0.567 (0.566, 0.569)} \\
    SRNN & 0.639 (0.637, 0.641) &0.693 (0.690, 0.695) & 0.661 (0.656, 0.666) & 0.562 (0.561, 0.564) & 0.553 (0.552, 0.555) \\
    KVAE & 0.559 (0.549, 0.570) & 0.608 (0.589, 0.627) & 0.565 (0.544, 0.586) & 0.538 (0.531, 0.545) & 0.523 (0.517, 0.529) \\
    TimeGAN & 0.558 (0.558, 0.558) & 0.703 (0.703, 0.704) & 0.552 (0.530, 0.573) & 0.545 (0.543, 0.548) & 0.530 (0.527, 0.533) \\
      \bottomrule
      \label{tab:synth_gen}
    \end{tabular}}
\end{table}

\subsection{Conditional generation}

From \Cref{tab:mimic_static} we see that many demographic variables have examples of highly underrepresented classes, possibly leading to the classification performance for these subpopulations to be much lower than for the majority class. This shines a light on a real problem found in many clinical settings, especially when transferring between hospitals \citep{covews}. 
In preliminary experiments, we investigate the classification score on a per-group level for the real data. While it does not occur for all static variables and classification tasks, we identify cases where there is a significant difference in the score of a given subpopulation, with respect to the other groups.
This inter-group performance gap raises the question if our model's conditional generation capability can be leveraged to address these fairness issues.

\begin{figure}[htp]
    \centering
    \includegraphics[width=0.9\textwidth]{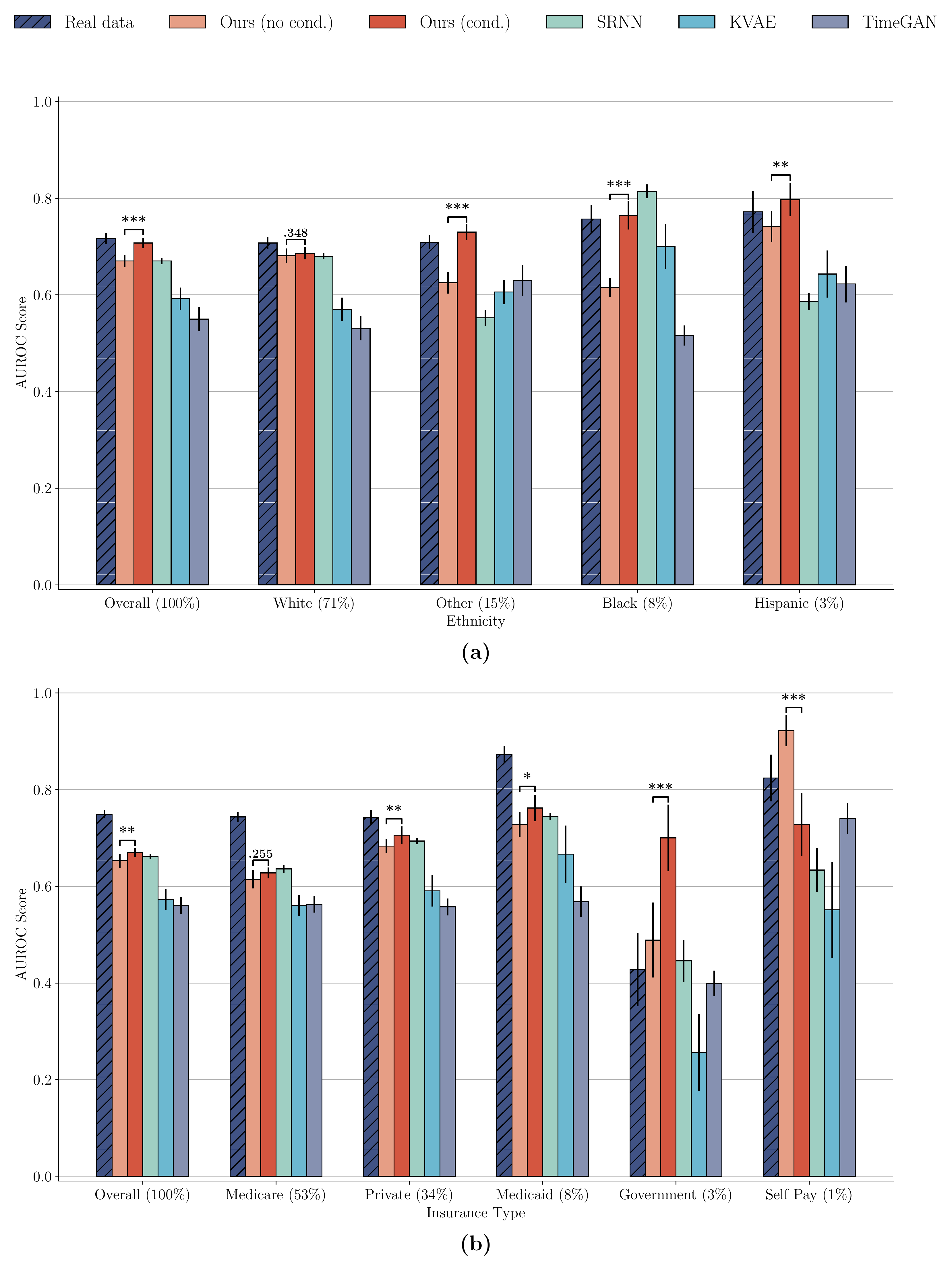}
    \caption{Comparison of AUROC score between our model when conditionally and unconditionally generating synthetic data with baselines. We show results of the \texttt{colloid\_bolus} task for different ethnicity groups (a) as wells as the range of insurance types (b). Note that the Asian subpopulation is not among the ethnicity groups, as there are no positive samples among this group for the considered task, thereby prohibiting the calculation of the AUROC score. Conditionally generating patient cohorts is favourable to unconditional generation overall, and for almost all subgroups, allowing for the generation of more representative and therefore fair synthetic data sets. Significance levels between groups of interest are shown with brackets, where * corresponds to p $<$ 0.05, ** to p $<$ 0.01 and *** to p $<$ 0.001}
    \label{fig:fair_synth}
\end{figure}

In the preceding experiment, we do no utilize our model's capacity to conditionally generate synthetic patient cohorts, as we do not explicitly condition our model on the static variables \(\*s\) during training or generation. 
Here, in addition to the label \(\*y\), we condition on a static variable of interest allowing us to then conditionally generate an equal number of synthetic patient samples for each subgroup of this considered static variable. We investigate if conditionally generating patient cohorts provides a benefit in terms of fairness, as well as overall downstream performance. In this experiment, we study the performance on a per-group level, comparing not only to the baseline approaches, but to our model when unconditionally generating the data, as well. To enable a fair comparison, the overall number of generated patients is equal for each considered model. The results of this experiment for two exemplary settings are presented in \Cref{fig:fair_synth}, with additional results reported in \Cref{app:cond_gen}.

The results indicate that, for settings in which the real data exhibits performance differences between subpopulations, conditonally generating synthetic patient cohorts provides a significant benefit over unconditional generation. 
In terms of overall performance, and in nearly all subgroups, our model outperforms the baselines when conditionally generating data. The variance of the scores across groups is also smaller when conditionally generating, than for any of the competing approaches.
Our model is not only capable of generating synthetic ``copies'' of the real data in terms of the distribution of demographics, but we can generate data sets with a high degree of control over the composition of subpopulations, resulting in more diverse training sets for downstream tasks.

\subsection{Real data augmentation}

In the preceding experiment, we demonstrated that our model's conditional generation capability can be used to synthesize patient cohorts that yield more fair downstream classification models. The settings that emerge in which our model can provide a benefit are those where the real data displays an imbalance in the downstream performance between patient subpopulations. This gives rise to the question if our approach to conditionally generate synthetic data can also be useful for the setting when access to the real data is not restricted, but rather the given cohort does not fulfill the requirements for the development of downstream models.
For these cases, we hypothesize that we can conditionally generate more examples of this previously underrepresented class, augment the real data with them and thereby boost the performance in the downstream classification task for this subpopulation.

\begin{figure}[htp]
    \centering
    \includegraphics[width=0.9\textwidth]{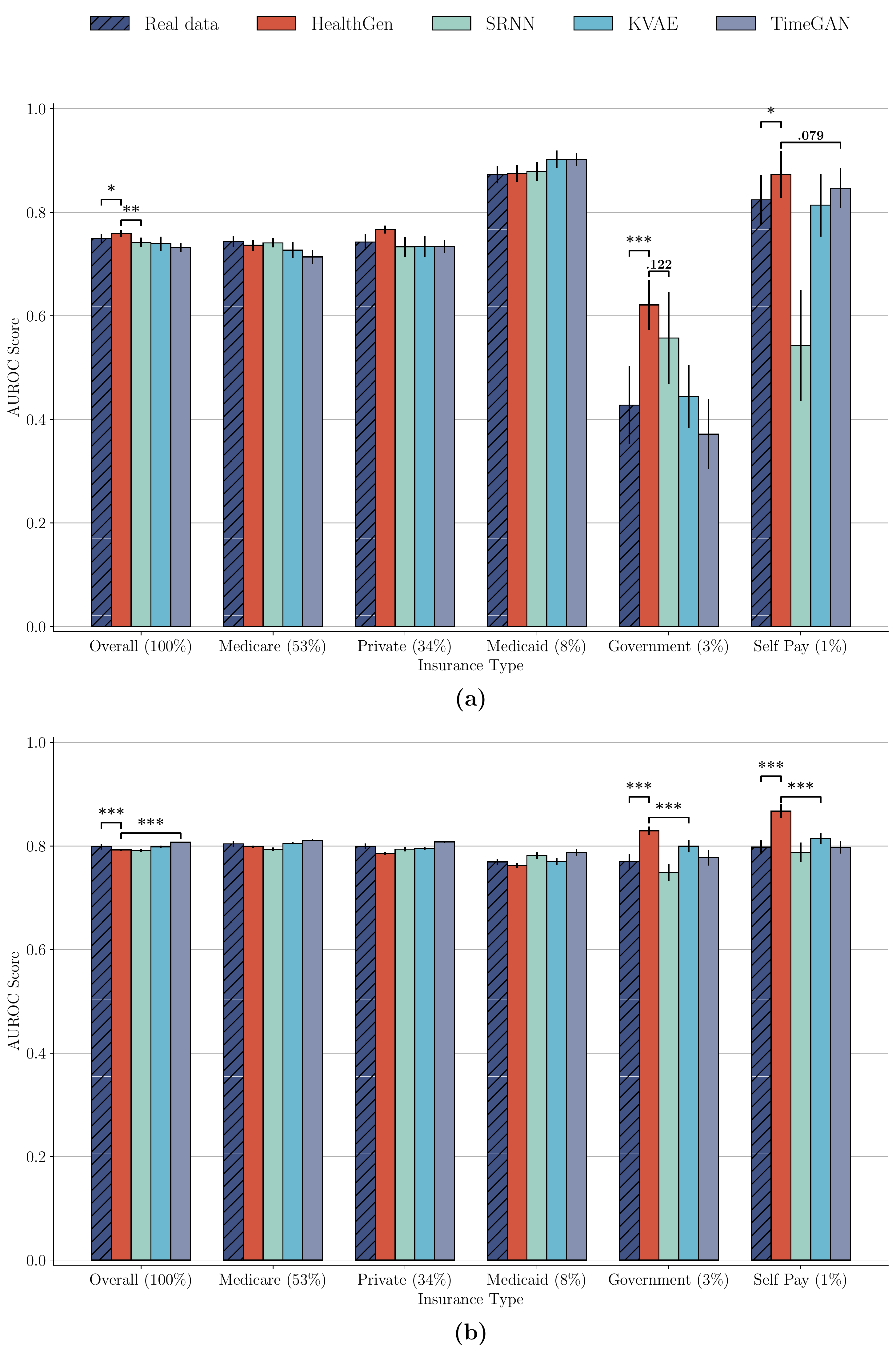}
    \caption{Comparison of the AUROC score between the real data and data sets obtained from the augmenting the real data with the synthetic patients from the considered generative models. We report the scores for each different insurance type on the (a) \texttt{colloid\_bolus} as well as the (b) \texttt{vaso} classification task. Significance levels between groups of interest are shown with brackets, where * corresponds to p $<$ 0.05, ** to p $<$ 0.01 and *** to p $<$ 0.001}
    \label{fig:fairness_aug}
\end{figure}

One of the cases, where we identify a significant difference in the performance of the trained classifier for different subtypes of patients, is the \texttt{colloid\_bolus} classification task when looking at different insurance types of patients. In \Cref{fig:fairness_aug}, we see that while the overall score on this task is fairly high, the underrepresented class of Government insured individuals has a significantly lower score than all other classes. The score of this class is even lower than 0.5, which would be obtained by randomly guessing which class a sample belongs to. 

To investigate if our model can improve the performance for such an underrepresented group, we conditionally generate additional samples of each underrepresented class and augment the real training data with these until all insurance types are fairly represented by the same number of samples. Since the baselines cannot generate data conditioned on static variables, we have them unconditionally generate the same number of overall samples that our model augments the real data with. We then compare the results of the downstream classifier trained on data sets augmented by synthetic data of the respective generative models to  the classifier trained on the real data. 

As we see in \Cref{fig:fairness_aug}(a), our model can indeed increase the performance of  previously underrepresented groups. Our model significantly boosts the predictive score of the Government class, without sacrificing the performance of any other subpopulation. Interestingly, the performance of the Self Pay class, which is also heavily underrepresented, is also boosted, even if it was already at a high level to begin with. While some other baselines also manage to boost the score of the Government insured class, they either fail to do so to the same degree as our approach, or they also decrease the performance for another class. Our model's superiority is further demonstrated by the fact that our conditional augmentation leads to a boosting of the overall score as well.

A second setting in which our model provides a benefit for underrepresented classes is the \texttt{vaso} task, again looking at different insurance types. Here, the performance on the minority groups of Government and Self Pay insured patients is not as dramatically lower compared to the other majority groups, but our approach to augment the real data still provides a significant benefit. Visualized in \Cref{fig:fairness_aug}(b), our augmentation boosts the performance of the downstream classifier for the two smallest classes significantly, even in a setting where their score is not severely below that of the larger groups to begin with.

\subsection{Privacy}

\begin{sidewaysfigure}[htp]
    \centering
    \includegraphics[width=\textwidth]{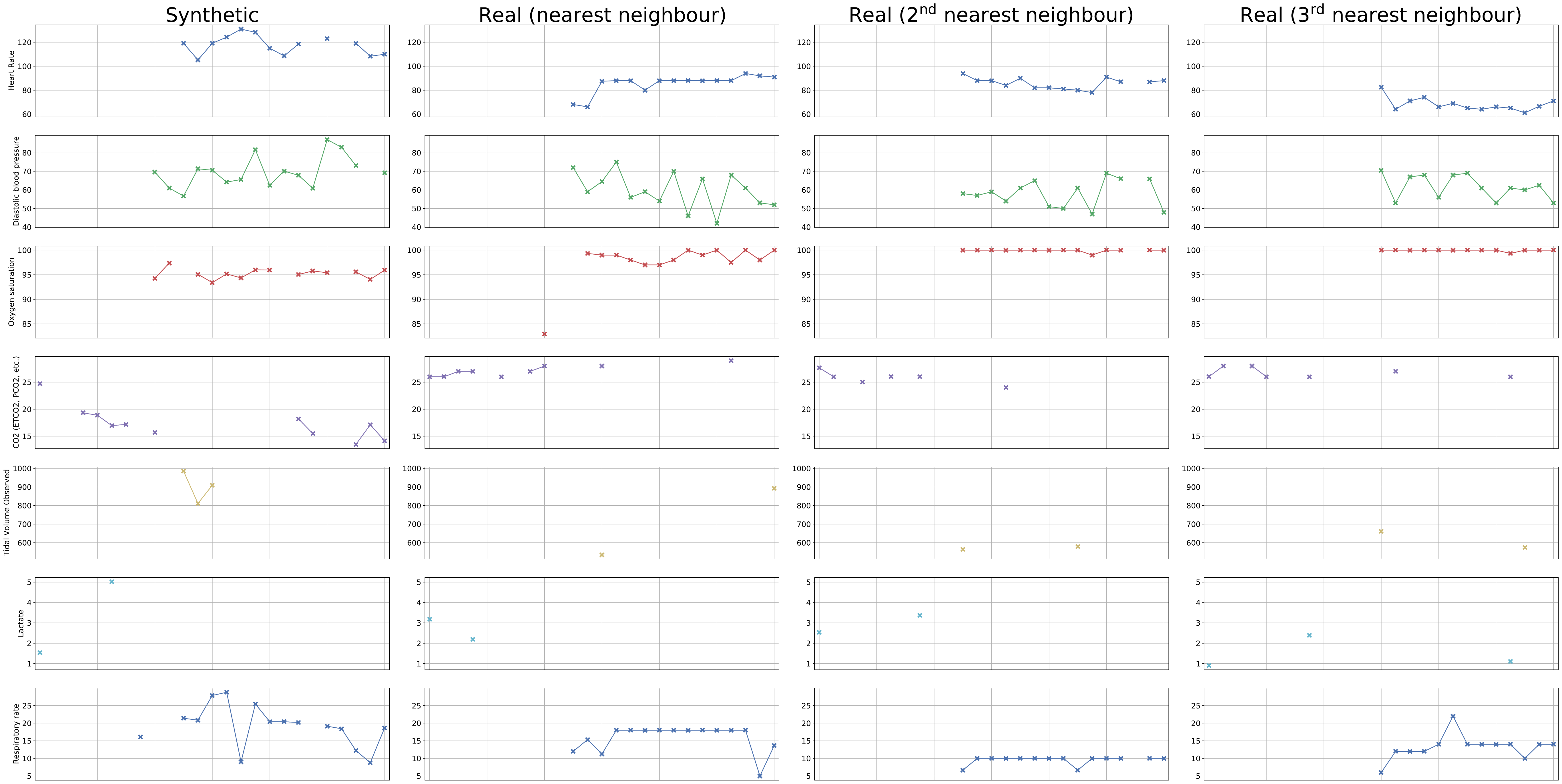}
    \caption{Comparison of the time series of a randomly sampled, synthetically generated patient and the corresponding time series of the three closest real patients (nearest neighbours) in the training data. While certain characteristics such as the number of missing values per feature or dynamics are similar between the synthetic sample and its nearest neighbours amongst the real data, we observe that the synthetically generated data is not a copy of the real data, indicating that our method does not memorize the data it sees during training.}
    \label{fig:privacy}
\end{sidewaysfigure}

To qualitatively assess if our model simply memorizes the training data and reproduces it at generation time, we visualize time series of a randomly selected, synthetically generated sample and time series of the three closest samples (nearest neighbours) in the training data. In \Cref{fig:privacy}, we compare the corresponding features of the synthetic and real samples side-by-side and observe that while certain patterns are shared, the synthetic data is not a copy of any of the real patients. This indicates that our model does not memorize the sensitive training data, allowing us to conjecture that it is privacy preserving to a certain degree, and sharing synthetically generated patient cohorts does not jeopardize the private information of the real patients our model was trained with.

\section{Discussion}
We presented HealthGen, a deep generative model capable of synthesizing realistic patient time series data, including informative patterns of missingness, that is faithful to data observed in real patient cohorts. To study the quality of the generated patient cohorts, we trained our generative model on the MIMIC-III data set, consisting of labeled ICU patient time series, to synthetically generate EHR data and evaluate the utility of the generated data on the clinically relevant downstream task of classifying patients' medical time series. In an experimental comparison against existing state-of-the-art models for time series generation, we explored multiple dimensions of the generative capability of our proposed approach: first, we synthesized patient cohorts with the same distribution of static variables as the real training data and observed that the data generated by our model is significantly more faithful to the real data across all evaluated downstream clinical prediction tasks than existing state-of-the-art approaches. In a second experiment, we demonstrated that HealthGen is capable of conditionally generating synthetic patient cohorts with static variable distributions that differ from the underlying, real data, without sacrificing the quality of the generated samples and boosting the fairness of the resulting synthetic patient cohorts. Finally, we identified settings where HealthGen can alleviate issues of unfair downstream classification performance between demographic subpopulations that arise in the real data, by augmenting the real patient cohorts with more diverse, synthetic samples.

\label{sec:disc}
\subsection{Generating synthetic patient cohorts}
A key motivation behind synthetically generating medical time series is the lack of access to this type of data for the development of downstream tasks. Data-driven approaches to assist clinical practitioners in diagnostic or therapeutic tasks promise significant improvements to the quality of healthcare we can provide in the future, but without sufficient data, both in terms of amount and quality, their development is impeded. Clinical institutions that collect this type of data at large are reluctant to centralize and share it, raising the question of how access to useful training and development data may be ensured. One approach that has been brought forward recently is the idea of synthesizing patient cohorts, with the hope that these generated data sets can then be shared freely. In this scenario, only one model hast to be granted access to the sensitive real data, while the synthesized cohorts that are generated by the trained generative model can be freely shared with anyone in need of data for developing a downstream task.
The primary requirement for this generated data is that it must adhere to the characteristics of the real data in such a way that it allows for the meaningful development of downstream models. These models are then deployed in the real world, to be used with real data. We demonstrate our model's capacity to fulfill precisely these requirements. On five different clinical downstream tasks, the synthetic patient cohorts generated by our approach are closer aligned to the real data, evident from their classification scores being closer to that of the downstream task trained on real data, than any competing baselines. While in no setting the generated data ever outperforms the real data in terms of classification score, we stress that this cannot be expected and more importantly this does not diminish the obtained results. In practice, one would only have access to synthetic data for development of these downstream models, so the performance obtained by training on real data is only considered for model selection of the upstream generative model, by means of providing a point of reference for comparison.

Our approach to synthetically generate realistic and useful medical time series data outperforms competing state-of-the-art models for a number of reasons. We include inductive biases aligned with the healthcare domain, in the form of explicitly modelling missing data patterns and separating the generation of these missing values from the generation of observable clinical variables. Furthermore, our model's capability to capture the influence of static and demographic variables on the generated data and to condition on them during the generation of the synthetic data adds to the expressive capacity of our architecture.
To the best of our knowledge, no other models to generate time series in the medical domain explicitly model missing data patterns, even though their importance and prevalence in clinical data is well known. Instead of cherry picking features with low missingness or downsampling the temporal resolution of the data to alleviate missing values, we can generate time series data that is faithful to the characteristics of realistically occurring EHR data.
Not only do we outperform the competing baselines on all of the considered downstream tasks, but we do so in a much more realistic setting than previously presented in the literature. This is a notable contribution, as we validate and compare models to generate medical time series with real-world downstream tasks for healthcare applications, giving our findings significantly more weight for clinical practitioners concerned with an impact beyond exemplary academic settings.

\subsection{Conditional generation}

In the context of healthcare applications, providing fair diagnostic models is of high ethical importance and increasing the fairness of such tools can have a potentially large impact. If an approach works well for the majority of a cohort at the cost of neglecting a certain subclass this can lead to systematically worse treatment of patients belonging to this group. Even a small increase in the predictive quality of a diagnostic tool can mean that hundreds or thousands of patients receive a treatment better aligned with their needs.

In addition to our model being able to generate synthetic patient cohorts of high quality and usefulness, we can do so with a high fidelity of control over the composition of cohorts, without having to sacrifice the quality of the generated data. In settings where the real data displays significant differences in performance between different subpopulations, utilizing this conditioning capability provides a benefit and yields more fair synthetic data sets. This increased fairness is evident from the lower variance between subpopulations when utilizing conditioning, compared to the unconditionally generated data of the other approaches. Importantly, this increase in fairness does not come at the cost of diminishing the performance of certain populations, but rather through an increase in the score of previously sub-par groups, which is also evident in the increase in overall score, with respect to our model when we do not condition.
While conditional generation never hurts overall, we cannot boost the score of any subpopulation in any arbitrary setting. For the conditioning to provide a benefit, the real data must display performance imbalances between subpopulations. When this imbalance is not present and all subgroups perform similarly, conditioning should not be expected to provide an additional benefit, as the differences in subpopulations are not relevant for their classification.

The fact that we can successfully condition the generative process of our model to synthesize patients with given features indicates HealthGen's capability to correctly capture meaningful dependencies between high-level, time invariant patient features and their influence on the resulting dynamics of the generated covariates. The key modelling choices that enable this level of conditioning are the fact that we introduce an additional static latent variable to capture time invariant patient states, as well as the inference procedure by which our model learns the dependencies between this time-invariant latent variable and the dynamics of the sequential data we are interested in generating. Splitting the high-level patient features from the dynamics on an architectural level encourages our model to focus on learning these concepts separately. Independence however does not follow from this separation, as high-level patient states will dictate the evolution of dynamical variables over time, which we capture in the dependencies of the static latent variable on time-varying observations during inference and vice versa during generation.

\subsection{Real data augmentation}
We have shown that by leveraging our model's conditional generation capability, we can synthesize data sets which are significantly more fair in their representation of subpopulations of patients. Having demonstrated this in the setting where access to the real data is not given for the development of downstream models, therefore having to rely on fully synthetic data, the question arises if conditionally generated data can also provide a benefit when we \textit{do} have access to the real data.
In a scenario where access to a real data set is given during the development of such a clinical tool, we investigated if augmenting the real data with synthetic patients of specific, underrepresented subpopulations can help to develop a more fair downstream classifier.

After identifying settings where certain subpopulations display significantly lower classification performances than the majority groups, in our final experiment, we demonstrate our model's capacity to increase the fairness of these real data sets through augmentation with synthetically generated data. This underlines the usefulness of synthetic data generation for a data augmentation task, which is orthogonal to the original objective of our model, namely generation of fully synthetic patient cohorts. That we can boost the performance of downstream tasks using these mixed-modality data sets consisting of real as well as synthetic data speaks to our model's capability to generate time series that are true to the real data in their informative characteristics and opens up even more possibilities of utilizing synthetically generated data in relevant, real-world applications.

Here, the effect of our approach's explicit modelling of static variables of interest and the resulting capability to condition on these becomes even more evident. While other generative models can also boost the classification of individual underrepresented classes through unconditional generation, our model proves to have a decisive advantage. The baselines are bound to generate some examples of the minority classes during generation, but we can generate these with high fidelity in a targeted fashion. The resulting augmented data set that follows from our approach manages to boost the score of underrepresented groups, without sacrificing the previously good score of any other subpopulations, which cannot be said for the baselines against which we compare.
While we can provide a benefit via augmentation with generated data in specific settings, this does not hold in general. This implies that we cannot simply hope to boost any subpopulations downstream classification performance by generating more samples of this class. Our findings suggest that two main conditions must be met for augmentation to provide a benefit: the gap between the classification score of the minority group and the other groups must exceed a certain magnitude and the other groups must display a minimum score overall, in order for the model to have informative enough examples to learn from, even if they belong to an other group than the one we are interested in generating.

\subsection{Limitations}
In this work, we do not provide any strong guarantees on the privacy preserving nature of the generated data sets. While it may be interesting to investigate and extend our model in the future in terms of rigorous differential privacy-preserving guarantees \citep{diff_priv}, we argue that we still generate synthetic data that is privacy-preserving to certain degree.
For example, we provide experimental evidence that our model does not simply memorize the training data and reproduce it to generate synthetic cohorts.
Moreover, it has also been shown that training neural network architectures with stochastic gradient descent intrinsically guarantees a certain degree of privacy \citep{hyland2020empirical}, the extent of which is however still an open research question.

Furthermore, the quality and diversity of the generated data which our model produces is limited by the real data with which it is trained. We cannot hope to generate samples of data which are too far out of the distribution of patients which the model has seen during training. A possible solution to this could be the integration of HealthGen into a federated learning framework as an avenue of future development. The initial motivation to synthetically generate EHR data is the lack of large publicly available data sets, with those that are available being only representative of a specific patient cohort. Training a generative model on multiple cohorts in a privacy preserving, federated fashion has been proposed to increase the diversity of the generated data and further catalyze the development of personalized medicine \citep{fed_future, fedlearn}.

\section{Methods}
\label{sec:methods}

\subsection{Data set and preprocessing}

In our experiments, we use the publicly available Medical Information Mart for Intensive Care (MIMIC-III) data set \citep{mimiciii} as input. In its raw form, it consists of the deidentified electronic health records (EHRs) of 53,423 patients collected in the intensive care units (ICUs) of the Beth Israel Deaconess Medical Center in Boston, Massachusetts, USA between 2001 and 2012. 
It consists of multiple tables containing the data of its over 50,000 patients. A single patient's information is linked across tables through a unique patient ID, and time series data contains a time stamp to maintain the correct temporal ordering of measurements. In this form, the sequential data is not ordered and many of the raw measurements represent the same concept, but are redundantly recorded under different names. 

As a first preprocessing step, we employ a slightly modified version of the \texttt{MIMIC-Extract} pipeline \citep{mimic_extract}. This yields a data set containing the ordered time series of measurements of each patient, static patient variables such as age, sex, ethnicity and insurance type, and a sequence of binary labels at each time step, indicating whether a certain medical intervention was active or not. We apply the standard cohort selection found in the literature \citep{mimic_use_1, mimic_use_2, mimic_use_3}: the first ICU admission of adult patients (at least 15 years old), with a minimum duration of at least 12 hours, resulting in a total number of \(N = 34472\) patients.

At this point, the time series data is still irregularly sampled and asynchronous across different features of the same patient.
Given a sampling frequency, we look at the resulting window around each time step and either record the measurement, or indicate the absence of a measurement with a \texttt{NaN} value. We then truncate all time series to have the same, fixed length. In our setting we choose a sampling frequency of 4 steps per hour and truncate the sequences to have a total duration of 12 hours.

From the observed feature sequences we additionally extract a sequence of binary masks \(\*m_{1:T}\) indicating where a value in \(\*x_{1:T}\) is missing:
\begin{equation}
    m_{t,d} = 
    \begin{cases}
      1, & \text{if } x_{t,d} \neq \texttt{NaN}, \\
      0, & \text{otherwise}.
    \end{cases}
\end{equation}
Finally, we standardize all (non-missing) numerical values of \(\*x_{1:T}\) to empirically have zero mean and unit variance along each dimension \(d \in D\), and replace the \texttt{NaN} values in \(\*x_{1:T}\) with zeros.

To obtain a binary label for a patient, we split the 12-hour sequence into three sections: a 6-hour observation window followed by a 2-hour hold-out section and finally a 4-hour prediction window. The label is then extracted from the prediction window: if an intervention is active at any time in this section, the label is positive, otherwise it is negative. Drawing inspiration from \citet{suresh2017clinical}, this procedure aims to create a fair prediction of future interventions from observed data by minimizing information leakage. If there is no gap between observation and prediction, oftentimes the last step of the observation contains enough information for a meaningful prediction.
We extract five binary labels corresponding to different types of clinical interventions in the ICU: \texttt{vent} refers to mechanical ventilation, \texttt{vaso} to the administration of vasopressor drugs, \texttt{colloid\_bolus} and \texttt{crystalloid\_bolus} refer to colloid and crystalloid fluid bolus administration, respectively, and \texttt{niv} denotes non-invasive ventilation. An overview of the prevalence of overall positive samples for each of these labels is presented in \Cref{tab:mimic_labels}.
\Cref{tab:mimic_static} provides a summary of the extracted static variables and the representation of each sub-cohort and \Cref{tab:mimic_stats} presents all extracted time-varying features together with selected statistics.

After preprocessing, each patient is represented by a time series of inputs \(\*x_{1:T} = \{\*x_t \in \mathbb{R}^D\}_{t=1}^T\), a time series of missingness masks \(\*m_{1:T} = \{\*m_t \in \{0,1\}^D\}_{t=1}^T\), where \(D = 104\), a vector of static features \(\*s \in \mathbb{R}^M, M = 4\) and a set of binary outcome labels \(\*y \in \{0, 1\}^L, L = 5\). The time series \(\*x_{1:T}\) and \(\*m_{1:T}\) cover 6 hours of measurements at a resolution of 15 minutes between steps, resulting in a sequence of length 25.
The final data set \(\mathcal{D} = \{\*x^n_{1:T}, \*m^n_{1:T}, \*s^n, \*y^n\}^N_{n=1}\) is then split into a training, validation and test set, stratified with respect to the labels~\(\*y\).

In \Cref{fig:mimic_examples} we visualize an exemplary set of time series of one patient. We can observe that some covariates such as the heart rate or oxygen saturation have many successive measurements and their evolution over time can be directly studied, while others such as CO2 are missing values at the majority of the time steps and the signal of their dynamics is much sparser. This visualization also shows the two types of correlations evident in the  sequential data. Firstly, values of variables may be correlated over time, as we can see from the evolutions of the heart rate and the diastolic blood pressure. Secondly, \textit{when} values were measured may correlate as well, i.e. the patterns of missingness for different input variables can be related.

\subsection{The HealthGen Model}

Here we introduce the main technical contribution of this work: the generative model we propose for the task of conditionally generating medical time series data, which we christen \textbf{HealthGen}. The HealthGen model consists of a dynamical VAE-based architecture that allows for the generation of feature time series with informative missing values, conditioned on high-level static variables and binary labels. 

As discussed in the previous section, our data consists of a feature time series \(\*x_{1:T}\) representing the physiological state of a patient, a sequence of binary missingness masks \(\*m_{1:T}\) indicating when a value of \(\*x_{1:T}\) is observed and when it is missing, static observable variables $\*s$, and labels $\*y$.
The pattern of missingness is notably not random, and highly informative, as preliminary experiments have shown (see \Cref{app:results}). This result is in line with the findings of \citet{grud}, who show that missing values in medical time series play a key role in downstream predictive tasks. 
The high informativeness of missingness patterns can be explained by imagining a patient with a visibly deteriorating state: clinical staff are much more likely to increase the number and frequency of measurements in this case, compared to a patient whose state is stable.
Given their evident importance, we explicitly model the missingness masks \(\*m_{1:T}\) alongside the observed feature sequence \(\*x_{1:T}\).

\paragraph{Generative model.}
The generative process starts from the static latent variable \(\*v\) with a fixed unconditional prior \(p(\*v) = \mathcal{N}(\*v; \*0, \*I)\). The observed features and missingness masks are then independently generated as follows.
The sequence of missingness masks \(\*m_{1:T}\) is generated from the static latent variable \(\*v\), conditioned on the static observable variable \(\*s\) and label \(\*y\). We do not model this as a dynamical process, but rather generate the entire sequence in one step, and model it with independent Bernoulli distributions:
\begin{align}
    \pthetam(\*m_{1:T} | \*v, \*s, \*y) &= \prod_{t=1}^T \prod_{d=1}^D \mathrm{Bernoulli}(m_{t,d}; \mu_{t,d}),
\end{align}
where the matrix of probabilities $\boldsymbol{\mu} \in [0, 1]^{T\times D}$ is given by a DNN $d_{\*m}(\*v, \*s, \*y)$ with a sigmoid on the last layer.
\begin{figure}[t]
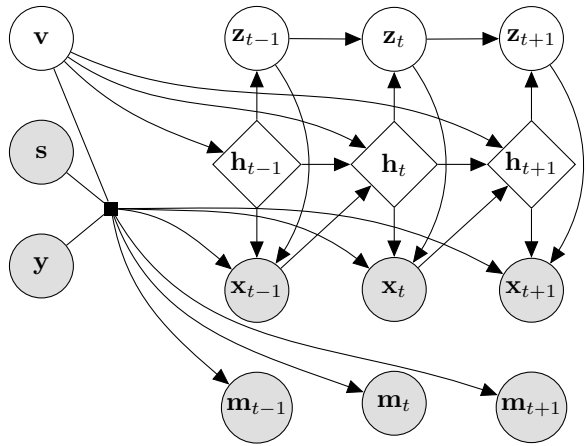

    \centering
    \tikz[latent/.append style={minimum size=0.85cm},obs/.append style={minimum size=0.85cm},det/.append style={minimum size=1.15cm}]{
        \node[det](h_tm1){\(\*h_{t-1}\)};
        \node[det, right=of h_tm1, xshift=-0.35cm, yshift=0.00cm](h_t){\(\*h_{t}\)};
        \node[det, right=of h_t, xshift=-0.35cm, yshift=0.00cm](h_tp1){\(\*h_{t+1}\)};
        \node[latent, above=of h_tm1, xshift=0.00cm, yshift=-0.35cm](z_tm1){\(\*z_{t-1}\)};
        \node[latent, above=of h_t, xshift=0.00cm, yshift=-0.35cm](z_t){\(\*z_{t}\)};
        \node[latent, above=of h_tp1, xshift=0.00cm, yshift=-0.35cm](z_tp1){\(\*z_{t+1}\)};
        \node[obs, below=of h_tm1, xshift=0.00cm, yshift=0.35cm](x_tm1){\(\*x_{t-1}\)};
        \node[obs, below=of h_t, xshift=0.00cm, yshift=0.35cm](x_t){\(\*x_{t}\)};
        \node[obs, below=of h_tp1, xshift=0.00cm, yshift=0.35cm](x_tp1){\(\*x_{t+1}\)};
        \node[obs, below=of x_tm1, xshift=0.00cm, yshift=0.35cm](m_tm1){\(\*m_{t-1}\)};
        \node[obs, below=of x_t, xshift=0.00cm, yshift=0.35cm](m_t){\(\*m_{t}\)};
        \node[obs, below=of x_tp1, xshift=0.00cm, yshift=0.35cm](m_tp1){\(\*m_{t+1}\)};
        \node[latent, left=of z_tm1, xshift=-1.0cm, yshift=0.00cm](v){\(\*v\)};
        \node[obs, below=of v, xshift=-0.00cm, yshift=0.35cm](s){\(\*s\)};
        \node[obs, below=of s, xshift=-0.00cm, yshift=0.35cm](y){\(\*y\)};
        \factor[right=of s, yshift=-0.75cm] {cat} {} {} {};
        \factoredge {v,s,y} {cat} {};
        \draw[->] (cat) to [out=-5, in=140] (x_tm1);
        \draw[->] (cat) to [out=-2, in=140] (x_t);
        \draw[->] (cat) to [out=-0, in=150] (x_tp1);
        \draw[->] (cat) to [out=-80, in=140] (m_tm1);
        \draw[->] (cat) to [out=-70, in=160] (m_t);
        \draw[->] (cat) to [out=-60, in=160] (m_tp1);
        \draw[->] (v) to [out=-45, in=160] (h_tm1);
        \draw[->] (v) to [out=-35, in=150] (h_t);
        \draw[->] (v) to [out=-25, in=150] (h_tp1);
        \edge{z_tm1}{z_t};
        \edge{z_t}{z_tp1};
        \edge{h_tm1}{h_t};
        \edge{h_t}{h_tp1};
        \edge{h_tm1}{z_tm1};
        \edge{h_t}{z_t};
        \edge{h_tp1}{z_tp1};
        \edge{h_tm1}{x_tm1};
        \edge{h_t}{x_t};
        \edge{h_tp1}{x_tp1};
        \edge{x_tm1}{h_t};
        \edge{x_t}{h_tp1};
        \draw[->] (z_tm1) to [out=-55, in=55] (x_tm1);
        \draw[->] (z_t) to [out=-55, in=55] (x_t);
        \draw[->] (z_tp1) to [out=-55, in=55] (x_tp1);
    }
    \caption{Probabilistic graphical model of the generative process of HealthGen.}
    \label{fig:healthgen:gen}
\end{figure}
The generation of the observed features sequence \(\*x_{1:T}\) is based on the SRNN model \citep{srnn}, with additional conditioning on the static latent \(\*v\), the static features \(\*s\), and the labels \(\*y\), inspired by DSAE \citep{dsae} and the conditional VAE \citep{cvae}. The generative model for \(\*x_{1:T}\) is given by:
\begin{align}
    \*h_1 &= e_{\*h}(\*0, \*v, \*0), \\
    \pthetaz(\*z_1|\*h_1) &= \mathcal{N}(\*z_1; \pmb{\mu}_{\theta_{\*z}}(\*0, \*h_1), \text{diag}\{\pmb{\sigma}^2_{\theta_{\*z}}(\*0, \*h_1)\}),\\
    \*h_t &= e_{\*h}(\*x_{t-1}, \*v, \*h_{t-1}) \text{ for }t>1, \\
    \pthetaz(\*z_t|\*z_{t-1}, \*h_t) &= \mathcal{N}(\*z_t; \pmb{\mu}_{\theta_{\*z}}(\*z_{t-1}, \*h_t), \text{diag}\{\pmb{\sigma}^2_{\theta_{\*z}}(\*z_{t-1}, \*h_t)\}) \text{ for }t>1,\\
    \pthetax(\*x_t|\*z_t, \*h_t, \*v, \*s, \*y) &= \mathcal{N}(\*x_t; \pmb{\mu}_{\theta_{\*x}}(\*z_t, \*h_t, \*v, \*s, \*y), \text{diag}\{\pmb{\sigma}^2_{\theta_{\*x}}(\*z_t, \*h_t, \*v, \*s, \*y)\}).
\end{align}
Finally, the joint distribution of all variables, conditional on the observed static features and labels, is:
\begin{align}
    p(\*x_{1:T}, &\*m_{1:T}, \*z_{1:T}, \*h_{1:T}, \*v | \*s, \*y) \label{eq:healthgen:gen}\\
    =\ &p(\*v) \pthetam(\*m_{1:T} | \*v, \*s, \*y) \nonumber\\ 
    & \prod^T_{t=1} \pthetax(\*x_t|\*z_t, \*h_t, \*v, \*s, \*y) \pthetaz(\*z_t|\*z_{t-1}, \*h_t)  \pthetah(\*h_t|\*x_{t-1}, \*h_{t-1}, \*v),\nonumber
\end{align}
where, abusing notation, we defined:
\begin{align}
    \pthetaz(\*z_1|\*z_0, \*h_1) &= \pthetaz(\*z_1|\*h_1) \\
    \pthetah(\*h_1|\*x_0, \*h_0, \*v) &= \pthetah(\*h_1| \*v)
\end{align}
and the hidden states $\*h_t$ are deterministic:
\begin{align}
    &\pthetah(\*h_t|\*x_{t-1}, \*h_{t-1}, \*v) 
        = \delta(\*h_t - \tilde{\*h}_t) \\
    &\tilde{\*h}_t = 
        \begin{dcases} 
            e_{\*h}(\*0, \*v, \*0) &\text{ for } t=1\\
            e_{\*h}(\*x_{t-1}, \*v, \*h_{t-1}) &\text{ for }t>1
        \end{dcases}
\end{align}
The graphical model of \themethod{} is shown in \cref{fig:healthgen:gen}.

To generate synthetic data, we sample the features and missingness masks from the generative model $p(\*x_{1:T}, \*m_{1:T}, | \*s, \*y)$ using ancestral sampling as described in \Cref{alg:healthgen:sample}. In practice, the conditioning is implemented by concatenating the vectors \(\*s\) and \(\*y\), so to perform unconditional generation, we do not pass \(\*s\) at the conditioning step, but only \(\*y\). Note that we sample $\*z_0 \sim \mathcal{N}(\*z_0; \*0, \*I)$ rather than fixing it to $\*0$, as we observed that this empirically yields synthetic data that is more useful for the downstream tasks.

\begin{algorithm}[t]
    \caption{HealthGen sampling.}
    \label{alg:healthgen:sample}
    \begin{algorithmic} [1]
        \STATE Set values for conditionals \(\*s, \*y\)
        \STATE Sample \(\*v \sim p(\*v)\) from static latent prior
        \STATE Sample missingness masks \(\*m_{1:T} \sim \pthetam(\*m_{1:T} | \*v, \*s, \*y)\)
        \STATE Sample \(\*z_0 \sim \mathcal{N}(\*z_0; \*0, \*I)\)
        \STATE Initialize \(\*h_0 \leftarrow 0\)
        \FOR {\(t \leftarrow 1\) to \(T\)}
        \STATE Sample \(\*z_t \sim \pthetaz(\*z_t|\*z_{t-1}, \*h_t)\)
        \STATE Sample \(\*x_t \sim \pthetax(\*x_t|\*z_t, \*h_t, \*v, \*s, \*y)\)
        \STATE Encode \(\*x_t\) to obtain \(\*h_{t+1} \leftarrow e_{\*h}(\*x_t, \*v, \*h_t)\)
        \ENDFOR
        \RETURN \(\*x_{1:T}, \*m_{1:T}\)
    \end{algorithmic}
\end{algorithm}

\paragraph{Inference model.}
Similarly to the generative process, we split the inference model into two steps, beginning with the inference of the static variable \(\*v\) from the observable feature sequence, the missingness pattern sequence and the static features as well as the label. For the approximate posterior distribution of \(\*v\) we write:
\begin{equation}
    \qphiv(\*v | \*x_{1:T}, \*m_{1:T}, \*s, \*y) = \mathcal{N}(\*v; \pmb{\mu}_{\oldphi_{\*v}}(\*x_{1:T}, \*m_{1:T}, \*s, \*y), \text{diag}\{\pmb{\sigma}^2_{\oldphi_{\*v}}(\*x_{1:T}, \*m_{1:T}, \*s, \*y)\}).
\end{equation}
The static latent variable \(\*v\) encodes the static features as well as the label, but it also encodes static information from the time series inputs. This allows our model to capture high-level information about a patient's state, that is not explicitly represented by any of the static features alone. By splitting the inference into a static latent variable and a latent time series representing the underlying dynamics of the observable features, our model learns to separate the time invariant content of a given sample from the dynamics that govern the evolution of the time-varying parts of its state. A patient's general state has a large effect on the temporal evolution of their time-varying lower-level states, which is reflected in our model's conditioning on the static features (both latent and observed) at multiple steps during the inference and generative processes.

The approximate posterior of $\*z_{1:T}$ is defined as follows:
\begin{align}
    \*g_T &= e_{\*g}(\*x_T, \*h_T, \*v, \*0),\\
    \*g_t &= e_{\*g}(\*x_t, \*h_t, \*v, \*g_{t+1}) \text{ for }t<T,\\
    \qphiz(\*z_1 | \*g_1) &= \mathcal{N}(\*z_1; \pmb{\mu}_{\oldphi_{\*z}}(\*0, \*g_1), \text{diag}\{\pmb{\sigma}^2_{\oldphi_{\*z}}(\*0, \*g_1)\}),\\
    \qphiz(\*z_t | \*z_{t-1}, \*g_t) &= \mathcal{N}(\*z_t; \pmb{\mu}_{\oldphi_{\*z}}(\*z_{t-1}, \*g_t), \text{diag}\{\pmb{\sigma}^2_{\oldphi_{\*z}}(\*z_{t-1}, \*g_t)\}) \text{ for }t>1,
\end{align}
where \(e_{\*g}(.)\) is a backward RNN and \(\*h_t\) is the state of the forward RNN shared with the generative model. The overall inference model of \themethod{} can be then written as follows:
\begin{align}
    q_{\oldphi}(\*z_{1:T}, & \*g_{1:T}, \*h_{1:T}, \*v | \*x_{1:T}, \*m_{1:T}, \*s, \*y)\\
    =\ & \qphiv(\*v | \*x_{1:T}, \*m_{1:T}, \*s, \*y) \\
    & \prod^T_{t=1} \qphiz(\*z_t | \*z_{t-1}, \*g_t) \qphig(\*g_t | \*x_t, \*h_t, \*g_{t+1}, \*v) \pthetah(\*h_t|\*x_{t-1}, \*h_{t-1}, \*v),
\end{align}

\begin{figure}[t]
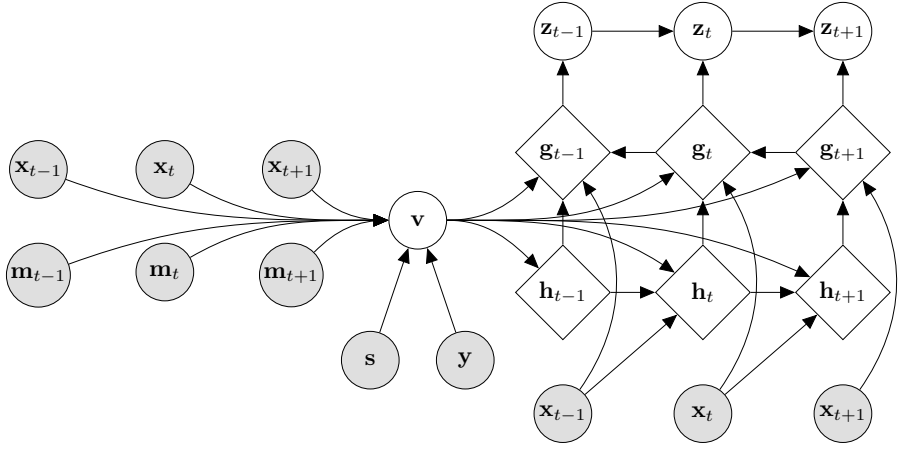

    \centering
    \resizebox{\textwidth}{!}{%
    \tikz[latent/.append style={minimum size=0.85cm},obs/.append style={minimum size=0.85cm},det/.append style={minimum size=1.4cm}]{
        \node[det](h_tm1){\(\*h_{t-1}\)};
        \node[det, right=of h_tm1, xshift=-0.35cm, yshift=0.00cm](h_t){\(\*h_{t}\)};
        \node[det, right=of h_t, xshift=-0.35cm, yshift=0.00cm](h_tp1){\(\*h_{t+1}\)};
        \node[det, above=of h_tm1, xshift=0.00cm, yshift=-0.35cm](g_tm1){\(\*g_{t-1}\)};
        \node[det, above=of h_t, xshift=0.00cm, yshift=-0.35cm](g_t){\(\*g_{t}\)};
        \node[det, above=of h_tp1, xshift=0.00cm, yshift=-0.35cm](g_tp1){\(\*g_{t+1}\)};
        \node[latent, above=of g_tm1, xshift=0.00cm, yshift=-0.35cm](z_tm1){\(\*z_{t-1}\)};
        \node[latent, above=of g_t, xshift=0.00cm, yshift=-0.35cm](z_t){\(\*z_{t}\)};
        \node[latent, above=of g_tp1, xshift=0.00cm, yshift=-0.35cm](z_tp1){\(\*z_{t+1}\)};
        \node[obs, below=of h_tm1, xshift=0.00cm, yshift=0.35cm](x_tm1){\(\*x_{t-1}\)};
        \node[obs, below=of h_t, xshift=0.00cm, yshift=0.35cm](x_t){\(\*x_{t}\)};
        \node[obs, below=of h_tp1, xshift=0.00cm, yshift=0.35cm](x_tp1){\(\*x_{t+1}\)};
        \node[latent, left=of g_tm1, xshift=-0.0cm, yshift=-1.00cm](v){\(\*v\)};
        \node[obs, left=of h_tm1, xshift=-0.70cm, yshift=-1.0cm](s){\(\*s\)};
        \node[obs, left=of h_tm1, xshift=0.70cm, yshift=-1.0cm](y){\(\*y\)};
        \node[obs, left=of v, xshift=-0.00cm, yshift=0.75cm](x_tp1_2){\(\*x_{t+1}\)};
        \node[obs, left=of x_tp1_2, xshift=-0.0cm, yshift=0.00cm](x_t_2){\(\*x_{t}\)};
        \node[obs, left=of x_t_2, xshift=-0.0cm, yshift=0.00cm](x_tm1_2){\(\*x_{t-1}\)};
        \node[obs, below=of x_tm1_2, xshift=0.00cm, yshift=0.35cm](m_tm1){\(\*m_{t-1}\)};
        \node[obs, below=of x_t_2, xshift=0.00cm, yshift=0.35cm](m_t){\(\*m_{t}\)};
        \node[obs, below=of x_tp1_2, xshift=0.00cm, yshift=0.35cm](m_tp1){\(\*m_{t+1}\)};
        \edge{h_tm1}{g_tm1, h_t};
        \edge{h_t}{g_t, h_tp1};
        \edge{h_tp1}{g_tp1};
        \edge{g_tp1}{g_t, z_tp1};
        \edge{g_t}{g_tm1, z_t};
        \edge{g_tm1}{z_tm1};
        \edge{z_tm1}{z_t};
        \edge{z_t}{z_tp1};
        \edge{x_tm1}{h_t};
        \edge{x_t}{h_tp1};
        \draw[->] (x_tm1) to [out=55, in=-55] (g_tm1);
        \draw[->] (x_t) to [out=55, in=-55] (g_t);
        \draw[->] (x_tp1) to [out=55, in=-55] (g_tp1);
        \draw[->] (v) to [out=0, in=-135] (g_tm1);
        \draw[->] (v) to [out=0, in=135] (h_tm1);
        \draw[->] (v) to [out=0, in=-145] (g_t);
        \draw[->] (v) to [out=0, in=145] (h_t);
        \draw[->] (v) to [out=0, in=-155] (g_tp1);
        \draw[->] (v) to [out=0, in=155] (h_tp1);
        \edge{s}{v};
        \edge{y}{v};
        \draw[->] (x_tp1_2) to [out=-45, in=-180] (v);
        \draw[->] (m_tp1) to [out=45, in=-180] (v);
        \draw[->] (x_t_2) to [out=-30, in=-180] (v);
        \draw[->] (m_t) to [out=30, in=-180] (v);
        \draw[->] (x_tm1_2) to [out=-20, in=-180] (v);
        \draw[->] (m_tm1) to [out=20, in=-180] (v);
    }
    }
    \caption{Probabilistic graphical model of HealthGen at inference time.}
    \label{fig:healthgen:inf}
\end{figure}
where, abusing notation, we defined:
\begin{align}
    \qphiz(\*z_1 | \*z_0, \*g_1) &= \qphiz(\*z_1 | \*g_1), \\
    \qphig(\*g_T | \*x_T, \*h_T, \*g_{T+1}, \*v) &= \qphig(\*g_T | \*x_T, \*h_T, \*v),
\end{align}
the hidden states $\*g_t$ are deterministic:
\begin{align}
    &\qphig(\*g_t | \*x_t, \*h_t, \*g_{t+1}, \*v) 
        = \delta(\*g_t - \tilde{\*g}_t) \\
    &\tilde{\*g}_t = 
        \begin{dcases} 
            e_{\*g}(\*x_t, \*h_t, \*v, \*0) &\text{ for } t=T,\\
            e_{\*g}(\*x_t, \*h_t, \*v, \*g_{t+1}) &\text{ for }t<T,
        \end{dcases}
\end{align}
and $\pthetah(\*h_t|\*x_{t-1}, \*h_{t-1}, \*v)$ was defined in the generative model.

The inference model of HealthGen is visualized in \Cref{fig:healthgen:inf}.

\paragraph{Training HealthGen.}
We optimize \themethod{} by maximizing the Evidence Lower BOund (ELBO), a lower bound to the data log likelihood conditional on the labels and observable static variables:
\begin{align}
    \log p(\*x, \*m | \*s, \*y) 
    &= \log \int_{\*z, \*v} p(\*x, \*m | \*s, \*y) d\*z  d\*v\\
    &= \log \E_{q(\*z, \*v | \*x, \*m, \*s, \*y)} \left[\dfrac{p(\*x, \*m, \*z, \*v| \*s, \*y)}{q(\*z,  \*v | \*x, \*m, \*s, \*y)}  \right]\\
    &\geq \E_{q(\*z, \*v | \*x, \*m, \*s, \*y)} \left[\log \dfrac{p(\*x, \*m, \*z, \*v| \*s, \*y)}{q(\*z, \*v | \*x, \*m, \*s, \*y)}  \right] 
    =: \mathcal{L}(\theta, \oldphi) \label{eq:elbo_healthgen}
\end{align}
where we dropped the subscript $1\!:\!T$ when referring to the entire sequence.
We can obtain the joint of the inference model in the denominator by marginalizing over $\*g$ and $\*h$, using the fact that their posteriors are deltas:
\begin{align}
    q(\*z, \*v | \*x, \*m, \*s, \*y) 
    &= \int_{\*g,\*h} q(\*z, \*h, \*g, \*v | \*x, \*m, \*s, \*y) d\*g d\*h \\
    &= \qphiv(\*v | \*x, \*m, \*s, \*y) \prod^T_{t=1} \qphiz(\*z_t | \*z_{t-1}, \tilde{\*g}_t),
\end{align}
where $\tilde{\*g}$ is the sequence of deterministic states of the backward RNN, and the states $\tilde{\*h}$ of the forward RNN are directly used only to compute $\tilde{\*g}$.
Similarly, we can marginalize $\*h$ in the generative model:
\begin{align*}
    p(\*x, \*m, \*z, \*v | \*s, \*y) 
    = p(\*v) \pthetam(\*m | \*v, \*s, \*y)
    \prod^T_{t=1} \pthetax(\*x_t|\*z_t, \tilde{\*h}_t, \*v, \*s, \*y) \pthetaz(\*z_t|\*z_{t-1}, \tilde{\*h}_t).
\end{align*}
The ELBO can finally be rewritten as follows:
\begin{align*}
    \mathcal{L}(\theta, \oldphi) 
    &= \E_{q(\*z, \*v | \*x, \*m, \*s, \*y)} \bigg[ 
        \log \pthetam(\*m | \*v, \*s, \*y)
        + \sum_t \log \pthetax(\*x_t | \*z_t, \tilde{\*h}_t, \*v, \*s, \*y)\\
        &\mspace{140mu} + \log \dfrac{p(\*v)}{\qphiv(\*v | \*x, \*m, \*s, \*y)}
        + \sum_t \log \dfrac{\pthetaz(\*z_t | \*z_{t-1}, \tilde{\*h}_t)}{\qphiz(\*z_t | \*z_{t-1}, \tilde{\*g}_t)}
    \bigg]\\
    &= \E_{\qphiv(\*v | \*x, \*m, \*s, \*y)} \bigg[
        \log \pthetam(\*m | \*v, \*s, \*y)  \\
        &\mspace{140mu} +\sum_t \E_{\qphiz(\*z_{1:t} | \tilde{\*g}_{1:t})} \left[ \log \pthetax(\*x_t | \*z_t, \tilde{\*h}_t, \*v, \*s, \*y)\right]\bigg]\\
        &\qquad - \KL(\qphiv(\*v | \*x, \*m, \*s, \*y) \| p(\*v)) \\
        &\qquad - \sum_t \E_{\qphiz(\*z_{1:t-1} | \tilde{\*g}_{1:t-1})} \left[\KL(\qphiz(\*z_t | \*z_{t-1}, \tilde{\*g}_t) \| \pthetaz(\*z_t | \*z_{t-1}, \tilde{\*h}_t))\right]
\end{align*}
The parameters \(\theta = [\theta_{\*x}, \theta_{\*m}, \theta_{\*z}, \theta_{\*h}]\) of the generative model and \(\oldphi = [\oldphi_{\*v}, \oldphi_{\*z}, \oldphi_{\*g}] \) of the inference model are jointly trained by descending on the negative gradient of \Cref{eq:elbo_healthgen}. The KL divergence terms have analytical expressions and all intractable expectations are approximated with Monte Carlo estimation. In practice, we mask the reconstruction loss term of the features \(\*x_{1:T}\) with the masks \(\*m_{1:T}\) to only take into account the learning signal of the features which have actually been observed.

\subsection{Baseline models}
\label{ssec:baselines}

We choose three generative models against which to compare our approach with: the SRNN \citep{srnn}, the KVAE \citep{kvae} and the TimeGAN \citep{timegan} models. These models were chosen in an effort to select examples from the literature which represent the state-of-the-art in generative sequence modelling for different architectures. For technical details on the baseline models, please refer to the original papers.

The SRNN model is chosen to represent the ``classical" dynamic VAE model: it utilizes RNNs as encoder and decoder and models the internal dynamics of the inferred latent sequence with an explicit transition model. In the comprehensive comparison between DVAE models provided by \citet{girin2020dynamical} it emerges as the most performative model, leading us to select it as the representative for this class of generative models.

The KalmanVAE is also included in our comparison due to its unique approach to model dynamics in the latent space. It combines a VAE with a classical linear state-space model to model the latent dynamics, resulting in interesting properties for the inference process. \citet{kvae} show that this approach works well in settings with well described dynamics, such as low dimensional mechanical systems, leading to the question of how well this translates to dynamics of clinical observables.

Models based on the GAN architecture have the reputation of shining when it comes to generating high quality synthetic data. To investigate if this is also the case in the setting we consider, we compare against the state-of-the-art GAN model for sequential data. In their original publication, \citet{timegan} also present one experiment on the MIMIC-III data set, making the TimeGAN model one of the most direct competitors to our approach a priori.

Since none of the models described above have the capability to generate data conditioned on labels \(\*y\), we provide a minor extension to all models, to enable a more fair comparison. Drawing inspiration from the Conditional VAE model \citep{cvae}, we repeat the labels \(\*y\) to all \(T\) time steps \(\*y_{1:T}\) and encode them as an additional feature during training. The resulting latent sequence is then again extended by \(\*y_{1:T}\) before decoding. At generation time, we can choose \(\*y_{1:T}\) as we wish, append it to the sampled prior or random noise and decode to obtain a generated sample conditioned on the label we desire.

\pgfdeclarelayer{background}
\pgfdeclarelayer{foreground}
\pgfsetlayers{background,main,foreground}
\begin{figure}[t]
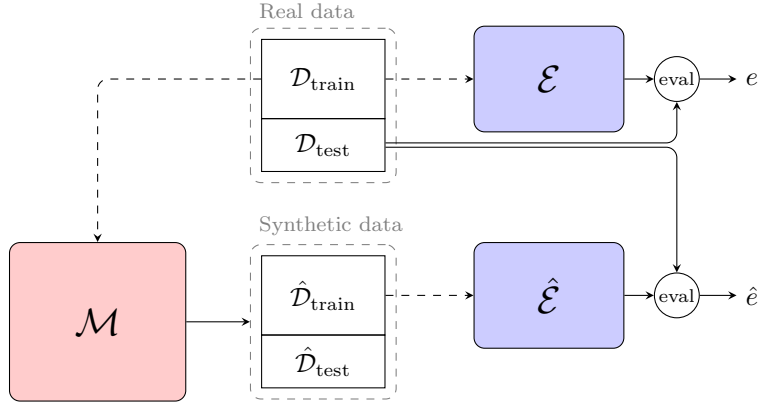

    \centering
    \tikz{
        \node (D_train) [draw, text width=4em, text centered, minimum height=3em] {\(\mathcal{D}_{\text{train}}\)};
        \node (D_test) [below=-0.4pt of D_train, draw, text width=4em, text centered, minimum height=2em] {\(\mathcal{D}_{\text{test}}\)};
        \node (D_hat_train) [below of=D_test, yshift=-1.0cm, draw, text width=4em, text centered, minimum height=3em] {\(\hat{\mathcal{D}}_{\text{train}}\)};
        \node (D_hat_test) [below=-0.4pt of D_hat_train, draw, text width=4em, text centered, minimum height=2em] {\(\hat{\mathcal{D}}_{\text{test}}\)};
        
        \begin{pgfonlayer}{background}
            \path (D_train.north west)+(0.1,-0.1) node (a) {};
            \path (D_test.south east)+(-0.1,0.1) node (b) {};
            \node[fit={(a) (b)}, rounded corners, draw=black!50, dashed, label={[opacity=0.5, shift=(real_data.north west)]above right:Real data}] (real_data) {};
            
            \path (D_hat_train.north west)+(0.1,-0.1) node (a) {};
            \path (D_hat_test.south east)+(-0.1,0.1) node (b) {};
            \node[fit={(a) (b)}, rounded corners, draw=black!50, dashed, label={[opacity=0.5, shift=(synth_data.north west)]above right:Synthetic data}] (synth_data) {};
        \end{pgfonlayer}
        
        \node (M) [left of=synth_data, xshift=-2cm, fill=red!20, draw, text width=6em, text centered, minimum height=6em, rounded corners, font=\Large] {\(\mathcal{M}\)};
        \draw [-stealth, rounded corners, dashed] (D_train) -| (M);
        \draw [-stealth] (M) -- (synth_data);
        
        \node (E_real) [right of=D_train, text width=5em, xshift=2cm, fill=blue!20, draw, text centered, minimum height=4em, rounded corners, font=\Large] {\(\mathcal{E}\)};
        \node (E_synth) [right of=D_hat_train, text width=5em, xshift=2cm, fill=blue!20, draw, text centered, minimum height=4em, rounded corners, font=\Large] {\(\hat{\mathcal{E}}\)};
        \node (eval_real) [circle,inner sep=1pt, minimum size=1pt, right of=E_real, xshift=0.7cm, draw, font=\scriptsize] {eval};
        \node (eval_synth) [circle,inner sep=1pt, minimum size=1pt, right of=E_synth, xshift=0.7cm, draw, font=\scriptsize] {eval};
        
        \draw [-stealth, dashed] (D_train) -- (E_real);
        \draw [-stealth, dashed] (D_hat_train) -- (E_synth);
        \draw [-stealth] (E_real) -- (eval_real);
        \draw [-stealth] (E_synth) -- (eval_synth);
        
        \draw [-stealth, rounded corners] (D_test.2.5) -| (eval_real);
        \draw [-stealth, rounded corners] (D_test.-2.5) -| (eval_synth);
        
        \node (score_real) [right of=eval_real, xshift=0.0cm] {\(e\)};
        \draw [-stealth] (eval_real) -- (score_real);
        
        \node (score_synth) [right of=eval_synth, xshift=0.0cm] {\(\hat{e}\)};
        \draw [-stealth] (eval_synth) -- (score_synth);
    }
    \caption{Conceptual overview of the experimental pipeline. The generative model \(\mathcal{M}\) is trained with the real training data, allowing it to generate a synthetic data set. Two identical evaluation models \(\mathcal{E}\) and \(\hat{\mathcal{E}}\) are then trained with the real or synthetic data, respectively. Finally, these evaluation models are tested on the real data, yielding the metric \(e\), derived from the real training set and that derived from the synthetic training data, \(\hat{e}\).}
    \label{fig:pipeline_overview}
\end{figure}

\subsection{Evaluation}
Quantitatively evaluating generative models is no trivial task, and in the setting where the generated data takes the form of real valued time series, this is especially true. Generative models that have been widely heralded as impressive examples of their class often convince the reader with generated human faces that are indiscernible from real images \citep{stylegan, nvae}. In the medical setting, where specialized domain knowledge is necessary to tell the difference between real and fake samples, the quality of generated imaging data is presented to clinical experts, who then discriminate between synthetic and real samples \citep{overcoming_barriers}. 

Unfortunately, none of these approaches apply to the medical time series data we aim to synthesize. It may be possible to identify generated data with extremely low quality by visual inspection, but after a certain fidelity is achieved, discerning between a``better" or ``worse" example of generated samples is no longer qualitatively possible.

To this end, we rely on the \textit{Train on Synthetic, Test on Real} (TSTR) evaluation paradigm, first introduced by \citet{rcgan}. A conceptual overview of our employed evaluation pipeline is presented in \Cref{fig:pipeline_overview}. Let \(\mathcal{E}\) denote the evaluation model trained on the real data \(\mathcal{D}_{\text{train}}\) and \(\hat{\mathcal{E}}\) denote the evaluation model trained on the synthetic data \(\hat{\mathcal{D}}_{\text{train}}\). \(\mathcal{E}\) and \(\hat{\mathcal{E}}\) share the same architecture and are trained according to identical procedures with the same hyperparameters. Both models are then evaluated on the \textit{same} held-out real test data \(\mathcal{D}_{\text{test}}\):
\begin{align}
    e & = \mathcal{E}(\mathcal{D}_{\text{test}}), \\
    \hat{e} & = \hat{\mathcal{E}}(\mathcal{D}_{\text{test}}).
\end{align}
The quantitative measure for how well a given generative model \(\mathcal{M}\) performs is then measured in the difference between \(e\) and \(\hat{e}\). If the generative model \(\mathcal{M}\) captures the dependencies in the real data that are informative for the downstream task represented by the evaluation model, and it is successful in synthesizing these in the generated data set \(\hat{\mathcal{D}}\), this is reflected in a score \(\hat{e}\) that is close, or ideally equal, to \(e\).

The model that implements \(\mathcal{E}\) in practice is the GRU-D model \citep{grud} for medical time series classification. Based on the Gated Recurrent Unit (GRU) \citep{gru}, this model was specifically introduced for classifying time series with missing values in the medical domain. It identifies two characteristics of missing values in the healthcare setting: first, the value of the missing variable tends to decay to some default value if its last measurement happened a long time ago (homeostasis), and second, the importance of an input variable will decrease if it has not been observed for an extended period of time.

These principles are modelled with two separate decay mechanisms. If a variable is missing for a number of time steps, its value decays toward the empirical mean of its measurements over time. The second decay is applied to the internal hidden state of the RNN cell, to model the waning importance of states that have not been updated in a while. In addition to the input features \(\*x_{1:T}\), the GRU-D model also takes the masks \(\*m_{1:T}\) as well as the time since the last measurement \(\pmb{\delta}_{1:T} = \{\pmb{\delta}_t \in \mathbb{R}^D\}^T_{t = 1}\) as direct input. 

While previous works have also used the TSTR framework to evaluate the quality of their generated medical time series data \citep{rcgan,  timegan}, we argue that our setting is better suited to evaluate generative models in the healthcare domain. The key difference we wish to highlight is the proxy task, implemented by the downstream evaluation model, that is chosen for the evaluation. Past approaches have attempted to predict the value of the next step in the input sequence \citep{timegan}, or to predict whether a time series surpasses a pre-defined threshold \citep{rcgan}. We opt for a downstream model that is specifically designed for a clinically relevant prediction task using real-world medical time series. This constitutes a setting much closer aligned to a real application in healthcare, and thus facilitating a comparison of generative models according to the relevant criteria instead of contrived metrics.

\subsection{Uncertainty estimation}

In all of our experiments, we repeat each run with five random initialisations of weights for the entire experimental pipeline, i.e. the generative model as well as the downstream evaluation model. For each generative model, we then choose the initialisation with the highest resulting downstream performance and estimate the 95\% confidence interval of the mean of the AUROC score by performing bootstrap resampling 30 times on the resulting generated synthetic data set. This allows us to report and compare not only the obtained performance of the models we consider, but also the uncertainty of our chosen metric.

In the second and final experiments of this work, we additionally perform statistical tests to quantify the significance levels between competing approaches. Here, we take the scores of the bootstrapping for settings we wish to compare and perform the one-sided, parametric-free, Mann-Whitney U test \citep{utest}, to determine the significance levels of competing approaches.

\subsection{Memorization analysis}

We analyze the privacy preserving characteristics of our model in similar fashion to \citet{overcoming_barriers}. To find the nearest neighbour of a synthetic sample, among the real data, we measure the distances between their respective latent encoding. To this end, we take our trained model and encode a randomly sampled synthetic patient, yielding a 32-dimensional static latent vector \(\*v\) and a 32-dimensional latent time series \(\*z_{1:T}\) with 25 time steps. After flattening the time dimension in \(\*z_{1:T}\) and concatenating the static latent vector \(\*v\), we end up with an 832-dimensional latent representation of the synthetic patient. We repeat this process for all patients in the real training data set, again yielding an 832-dimensional latent representation for each real patient. Then, utilizing the cosine distance measure between vectors, we find the three nearest neighbours of the randomly sampled synthetic patient and plot the respective time series of this generated patient and its nearest neighbours amongst the training data in order to qualitatively compare them. A randomly sampled synthetic patient with its three nearest neighbours is visualized in \Cref{fig:privacy}.

\section{Data availability}
\label{sec:data}

The utilized MIMIC-III data set \citep{mimiciii} is publicly available to researchers after having completed a course to certify their capability to handle sensitive patient data. The data may be requested at \url{https://physionet.org/content/mimiciii/1.4/}.

\section{Code availability}

The necessary code to reproduce our experimental results is available online at \url{https://github.com/simonbing/HealthGen} (MIT License).

%%%%%%%%%%%%%%%%%%%%%%%%%%%%%%%%%%%%%%%%%%%%
\clearpage

\bibliographystyle{plainnat}
\bibliography{bib.bib}

%%%%%%%%%%%%%%%%%%%%%%%%%%%%%%%%%%%%%%%%%%%%
\newpage
\appendix

\counterwithin{figure}{section}
\counterwithin{table}{section}

\section{Additional Results}
\label{app:results}
\subsection{On the importance of missing data}

As we can see from \Cref{tab:mimic_stats}, the time series in the MIMIC-III data set are riddled with missing values. In fact, more values in a given feature sequence are missing than present.  Naïvely, the first instinct may be to impute these missing entries with some value. Given missingness rates of over 90\% for most features, imputation becomes less of a viable option, but as previous works have shown, the patterns of missingness in medical time series data can be highly informative \citep{Razavian2015TemporalCN, grud}.

To investigate how much information is encapsulated by the missingness patterns alone, we conduct a preliminary experiment. Using the real data\footnote{In the preliminary experiments in this section, we use a reduced data set consisting of 50\% of the available patients and 50 out of 104 available features to enable faster iteration during development.}, we first train our downstream classification task on the full set of available inputs, that is the time series of the features \(\*x_{1:T}\) and the missingness patterns \(\*m_{1:T}\). We then train the same model using only the missingness masks \(\*m_{1:T}\) and compare their evaluation scores in \Cref{tab:real_miss}.

Evidently, the missing value patterns are highly informative for the considered downstream classification task, as AUROC score obtained from training the model solely on the masks is only marginally below the case when the features \(\*x_{1:T}\) are added. Given the importance of the missingness masks to predict later interventions, modelling their generation deserves special attention. 

\begin{table}[h]
    \centering
    \caption{AUROC scores for the \texttt{vent} classification of real data, using all features or only the missingness masks.}
    \begin{tabular}{ll}
         \toprule
         Input & AUROC \\
         \midrule
         Real data (feats \& masks) & 0.811 (0.809, 0.813) \\
         Real data (masks only) & 0.796 (0.794, 0.797) \\
         \bottomrule
    \end{tabular}
    \label{tab:real_miss}
\end{table}

\subsection{Modelling missingness patterns}
As the preliminary experiments of the preceding subsection show, the missingness patterns in the data are highly informative, meriting a deeper effort in modelling them. To this end, we compare multiple approaches to model the generative distribution of the missingess patterns \(\*m_{1:T}\).

To experimentally compare different methods to model the missing data, we train generative models with varying architectures using only the masks \(\*m_{1:T}\) as input and evaluate the usefulness of their respectively generated data for the \texttt{vent} classification task, according to the TSTR framework described in the Methods section.

A first natural choice is to model the time series of missing values \(\*m_{1:T}\) as a dynamical process, given their sequential nature. We compare two approaches to model the missingness patterns as a dynamical process: first, the SRNN model \citep{srnn}, which represents the prototypical dynamical VAE architecture using RNNs as encoder and decoder, and second the KVAE \citep{kvae}, which represents a dynamical VAE approach using an internal linear state-space model to model the dynamics of the data.

Instead of modelling the missingness patterns dynamically, we also test approaches to model them as sequence-level features. That is, instead of viewing \(\*m_{1:T}\) as a time series that is generated in a recurrent fashion, we model them as static features that are generated in a single step. We implement this idea using the vanilla VAE architecture \citep{vae}, where once we choose a Multi-Layer Perceptron (MLP) for the encoder and decoder, and for comparison a 1-D convolutional network to encode and decode.

We report the results of the experiments for all four architectures in \Cref{tab:miss_arch} and can see a clear trend when comparing their performance. The first observation to be made is that modelling the missingness patterns as sequence-level static features instead of being generated by a dynamical process yields significantly better results. The second conclusion we can draw from comparing the considered approaches is that the MLP architecture seems better suited to model the missing data. As a follow up experiment, we additionally encode the feature time series \(\*x_{1:T}\) together with \(\*m_{1:T}\), to a single latent variable, and generate both from this point in the latent space. In this setting the MLP again outperforms the convolutional architecture, leading us to adapt this approach and yielding the final model architecture which we present in the Methods section.

\begin{table}[b]
    \centering
    \caption{Comparison of different architectures to model missingness patterns.}
    \begin{tabular}{llc}
        \toprule
        Architecture & Input & AUROC \\
        \midrule
        MLP & masks & \textbf{0.756 (0.754, 0.758)} \\
        Conv. & masks & 0.722 (0.719, 0.725) \\
        SRNN & masks & 0.679  (0.676, 0.682) \\
        KVAE & masks & 0.515  (0.490, 0.539) \\
        \midrule
        MLP & feats \& masks & \textbf{0.780 (0.775, 0.784)} \\
        Conv. & feats \& masks & 0.770 (0.766, 0.775) \\
        \bottomrule
    \end{tabular}
    \label{tab:miss_arch}
\end{table}

Instead of naïvely treating the missingness patterns \(\*m_{1:T}\) and the observable features \(\*x_{1:T}\) identically and modelling them according to the same dynamical process, as the baseline models do, separating their generative processes is a key factor in generating realistic medical time series. Since two separate underlying mechanisms --- the decision when to perform a measurement and the evolution of the patient's physiological state --- give rise to the different time series, modelling them with separate generative processes is a natural choice. While these mechanisms are distinct for each type of features, they are not independent from each other, which we capture in the connections between inferred static and dynamic latent variables in our model's architecture. Interestingly, modelling the generation of the missingness patterns via a single latent variable, instead of as a dynamical process, works better in practice. This gives the impression that the state-space models we tested are ill-suited to model the dynamics of binary variables like the missingness patterns, and perhaps specialized dynamical models would fare better.

\subsection{Conditional generation}
\label{app:cond_gen}

\begin{figure}
    \centering
    \includegraphics[width=\textwidth]{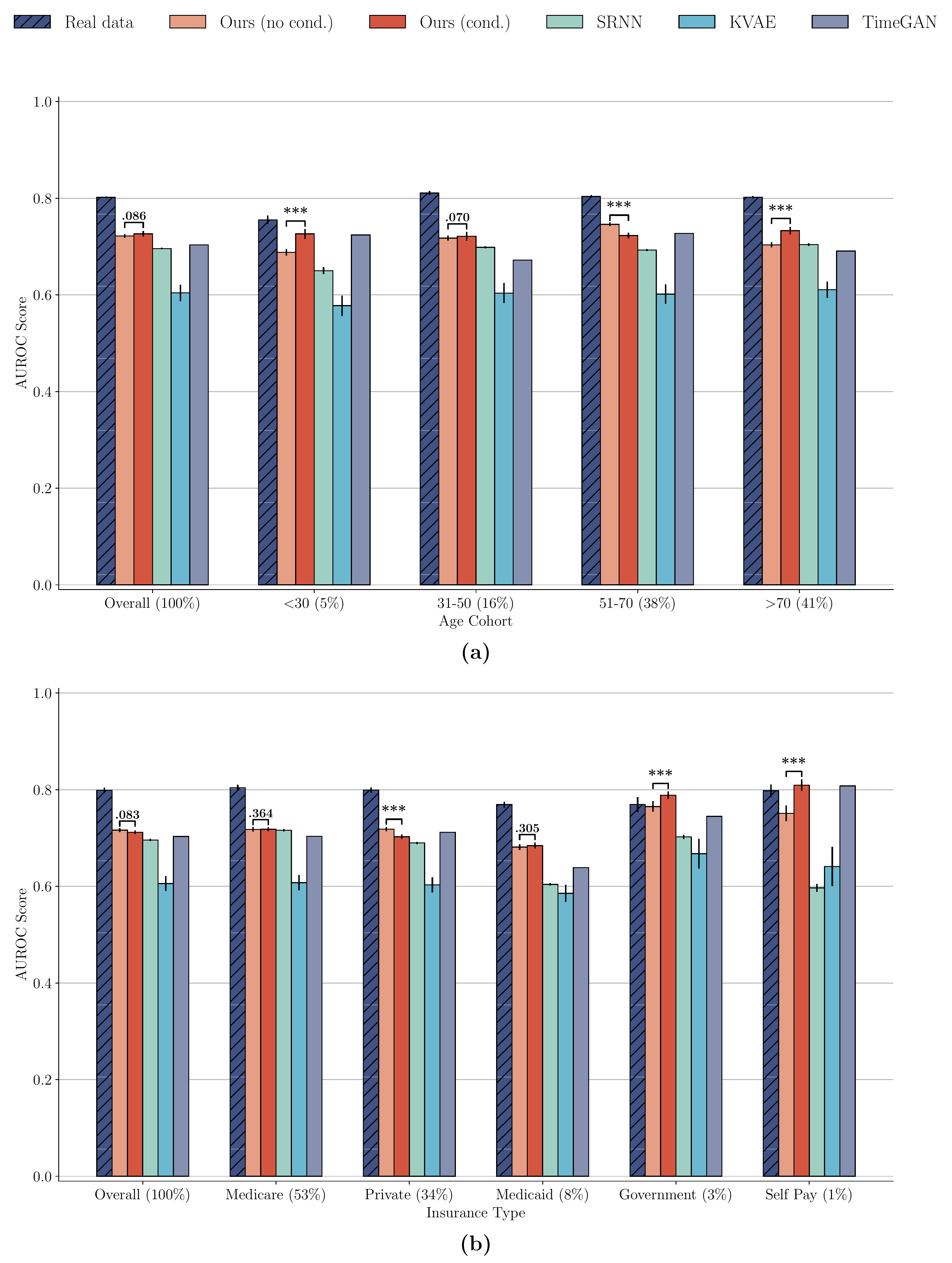}
    \caption{Comparison of AUROC score between our model when conditionally and unconditionally generating synthetic data with baselines. We show these additional results of the \texttt{vaso} task for different age cohorts (a) as well for the range of insurance types (b). Significance levels between groups of interest are shown with brackets, where * corresponds to p $<$ 0.05, ** to p $<$ 0.01 and *** to p $<$ 0.001}
    \label{fig:app_fair_synth}
\end{figure}

Here, we present additional settings of our conditional generation experiment, in which we utilize our model's capability to conditionally generate patient cohorts to yield data sets with an equal representation of subpopulations, ultimately increasing these data sets' fairness. These additional results are visualized in \cref{fig:app_fair_synth}.

\clearpage
\section{Implementation and Training Details}

Here we provide additional details on the neural architectures of our model, and on the training hyperparameters for all models.

The approximate posterior distribution $\qphiv(\*v | \*x_{1:T}, \*m_{1:T}, \*s, \*y)$ of the static latent variable \(\*v\) is parameterized by a 2-layer Multilayer Perceptron (MLP), as shown in \Cref{tab:enc_v}.
Both RNNs for \(\*h\) and \(\*g\) are implemented as GRU cells with an input size of 64 and hidden dimension 128. Their inputs are given by linear transformations of \([\*x_t, \*v]\) and \([\*x_t, \*h_t, \*v]\), respectively, followed by tanh.
The dynamics models of the latent variable \(\*z_t\), both at inference and generation time, are implemented as 2-layer MLPs, presented in \Cref{tab:dyn_z}.
The generation of feature vectors \(\*x_t\) at each time step as well as the full sequence of missingness masks \(\*m_{1:T}\) are also implemented as MLPs, with details provided in \Cref{tab:dec_x} and \Cref{tab:dec_m}, respectively.

We use the Adam optimization algorithm \citep{adam} to update the networks of all models presented in this work. The choices of hyperparameters for each model can be found in \Cref{tab:hyperparameters}. For the dynamical VAE baselines (SRNN and KVAE), we largely used the implementations provided by \citet{girin2020dynamical}, with some slight changes and extensions. The implementation of our model, as well as the experimental pipeline necessary to reproduce our results, can be found at \url{https://github.com/simonbing/HealthGen}.

Our experiments were conducted on a high-performance compute cluster, consisting mainly of NVIDIA V100 GPUs. In order to reproduce our results on a comparable hardware architecture, it would take approximately 200 GPU days.

\clearpage

\begin{table}
    \centering
    \caption{MLP encoder that parameterizes the approximate posterior distribution $\qphiv(\*v | \*x_{1:T}, \*m_{1:T}, \*s, \*y)$ of the static latent variable \(\*v\). The input is the concatenation of \([\*x_{1:T}, \*m_{1:T}, \*s, \*y]\) and the output vector is the mean and log variance of the distribution.}
    \begin{tabular}{cccc}
    \toprule
        Input size & Output size & Activation & Bias \\
        \midrule
        5201 & 256 & tanh & True \\
        256 & 128 & tanh & True \\
        128 & 64 & - & True \\
        \bottomrule
    \end{tabular}
    \label{tab:enc_v}
\end{table}

\begin{table}
    \centering
    \caption{MLP network architecture to parameterize \(q_{\oldphi_{\*z}}(\*z_t | \*z_{t-1}, \*g_t)\) and \(p_{\theta_{\*z}}(\*z_t | \*z_{t-1}, \*h_t)\), i.e. the latent dynamics in the inference and generative model. The inputs are the concatenation of \([\*z_{t-1}, \*g_t]\) for inference, and \([\*z_{t-1}, \*h_t]\) for generation. The output vector is the mean and log variance of the distribution.}
    \begin{tabular}{cccc}
    \toprule
        Input size & Output size & Activation & Bias \\
        \midrule
        160 & 64 & tanh & True \\
        64 & 32 & tanh & True \\
        32 & 64 & - & True \\
        \bottomrule
    \end{tabular}
    \label{tab:dyn_z}
\end{table}

\begin{table}
    \centering
    \caption{MLP decoder to parameterize the generative distribution of the features, \(p_{\theta_{\*x}}(\*x_t | \*z_t, \*h_t, \*v, \*s, \*y)\). The input is the concatenation of \([\*z_t, \*h_t, \*v, \*s, \*y]\), the output is the mean and log variance of the distribution.}
    \begin{tabular}{cccc}
    \toprule
        Input size & Output size & Activation & Bias \\
        \midrule
        193 & 256 & tanh & True \\
        256 & 208 & - & True \\
        \bottomrule
    \end{tabular}
    \label{tab:dec_x}
\end{table}

\begin{table}
    \centering
    \caption{MLP decoder for the generative distribution of the missingness masks, \(p_{\theta_{\*m}}(\*m_{1:T} | \*v, \*s, \*y)\). The input vector is the concatenation of \([\*v, \*s, \*y]\) and the output vector contains the mean \(\mu_{t,d}\) of the Bernoulli distributions that \(p_{\theta_{\*m}}(\*m_{1:T} | \*v, \*s, \*y)\) factorizes into.}
    \begin{tabular}{cccc}
    \toprule
        Input size & Output size & Activation & Bias \\
        \midrule
        33 & 128 & tanh & True \\
        128 & 256 & tanh & True \\
        256 & 2600 & sigmoid & True \\
        \bottomrule
    \end{tabular}
    \label{tab:dec_m}
\end{table}

\begin{table}
    \centering
    \caption{Hyperparameter choices for all of the considered models in this work. Hyperparameters were optimized by randomly sampling 20 different configurations for each model and then selecting the combination that yielded the highest AUROC score on the \texttt{vent} task.}
    \begin{tabular}{llr}
        \toprule
        Model & Hyperparameter & Range / \textbf{Value} \\
        \midrule
        \textbf{GRU-D} & Learning rate \(\alpha\) & \textbf{5e-4}, 1e-4, 1e-3 \\
        & Decay step for learning rate t [epochs] & 0, \textbf{20} \\
        & Dimension of hidden state \(\*h\) & 32, \textbf{64}, 128 \\
        \midrule
        \textbf{HealthGen} & Learning rate \(\alpha\) & \textbf{5e-4}, 1e-3, 5e-3 \\
        & Batch size & 32, \textbf{64}, 128 \\
        & Tradeoff parameter \(\beta\) & 0.5, 1.0, \textbf{5.0}, 10.0 \\
        & Dimension of static latent variable \(\*v\) & 16, \textbf{32}, 64 \\
        & Dimension of dynamic latent variable \(\*z\) & 16, \textbf{32}, 64 \\
        \midrule
        \textbf{SRNN} & Learning rate \(\alpha\) & \textbf{5e-4}, 1e-3, 5e-3 \\
        & Batch size & \textbf{32}, 64, 128 \\
        & Decay step for learning rate t [epochs] & 0, \textbf{20} \\
        & Dropout percentage \(p_{\text{dropout}}\) & \textbf{0}, 0.1, 0.2, 0.3 \\
        & Dimension of latent variable \(\*z\) & 8, \textbf{16}, 32 \\
        \midrule
        \textbf{KVAE} & Learning rate \(\alpha\) & 3e-6, 1e-5, \textbf{3e-5} \\
        & Batch size & 32, \textbf{64}, 128 \\
        & Decay step for learning rate t [epochs] & 0, \textbf{20} \\
        & Total training epochs & \textbf{40}, 50, 60, 70 \\
        & VAE only training epochs & \textbf{10}, 20, 30 \\
        & Kalman filter + VAE training epochs & 10, \textbf{20}, 30 \\
        & Tradeoff parameter \(\beta\) & 0.3, \textbf{0.5}, 1.0 \\
        & Number of linear state-space models \(K\) & 5, 10, \textbf{20} \\
        & Dimension of intermed. latent variable \(\*a\) & 16, \textbf{32} \\
        & Dimension of latent variable \(\*z\) & 8, \textbf{16} \\
        \midrule
        \textbf{TimeGAN} & Batch size & 32, \textbf{64}, 128 \\
        & Number of layers \(N\) & 2, 3, \textbf{5} \\
        & Dimension of the latent variable \(\*h\) & 8, 16, 24, \textbf{32} \\
        \bottomrule
    \end{tabular}
    \label{tab:hyperparameters}
\end{table}

\clearpage
\section{Data set characteristics}

\begin{small}
\begin{longtable}{lll}
        \caption{Descriptive statistics of the features included in the MIMIC-III data set. The Value column shows the median value of the respective variable with the 10th and 90th percentiles indicated in parantheses. The right-most column shows the percentage of missing values for each input feature.
        } \\
        \toprule
        \textbf{Feature} & \textbf{Value} & \textbf{Miss. [\%]} \\
        \midrule
        \endfirsthead
        \caption{(continued)} \\
        \toprule
        \textbf{Feature} & \textbf{Value} & \textbf{Miss. [\%]} \\
        \midrule 
        \endhead
        \bottomrule \endfoot
        Alanine aminotransferase & 31.00 (12.00, 216.95) & 98.198 \\
        Albumin & 3.30 (2.30, 4.20) & 98.710 \\
        Albumin ascites & 1.45 (0.78, 3.03) & 99.995 \\
        Albumin pleural & 2.00 (1.10, 2.60) & 99.996 \\
        Albumin urine & 13.20 (1.40, 161.19) & 99.994 \\
        Alkaline phosphate & 84.00 (48.00, 205.00) & 98.244 \\
        Anion gap & 14.00 (10.00, 20.00) & 94.942 \\
        Asparate aminotransferase & 41.00 (17.00, 321.50) & 98.197 \\
        Basophils & 0.30 (0.10, 0.85) & 98.533 \\
        Bicarbonate & 24.00 (18.00, 28.00) & 94.260 \\
        Bilirubin & 0.70 (0.30, 3.50) & 98.214 \\
        Blood urea nitrogen & 18.00 (9.00, 50.00) & 94.213 \\
        CO2 & 23.00 (17.00, 28.00) & 98.515 \\
        CO2 (ETCO2, PCO2, etc.) & 25.00 (20.00, 29.67) & 92.637 \\
        Calcium & 8.40 (7.20, 9.40) & 96.178 \\
        Calcium ionized & 1.14 (1.01, 1.28) & 95.743 \\
        Calcium urine & 2.40 (0.40, 13.08) & 99.994 \\
        Cardiac Index & 2.49 (1.82, 3.69) & 98.056 \\
        Cardiac Output Thermodilution & 4.90 (3.35, 7.20) & 98.299 \\
        Cardiac Output fick & 5.54 (3.70, 9.05) & 99.707 \\
        Central Venous Pressure & 11.00 (5.00, 17.00) & 91.292 \\
        Chloride & 105.00 (98.00, 112.00) & 92.977 \\
        Chloride urine & 47.00 (14.00, 116.60) & 99.902 \\
        Cholesterol & 164.00 (110.80, 226.20) & 99.796 \\
        Cholesterol HDL & 45.00 (30.00, 66.70) & 99.834 \\
        Cholesterol LDL & 93.00 (47.00, 147.00) & 99.843 \\
        Creatinine & 0.90 (0.60, 2.40) & 94.201 \\
        Creatinine ascites & 1.10 (0.56, 6.76) & 99.998 \\
        Creatinine body fluid & 1.10 (0.94, 1.66) & 99.999 \\
        Creatinine pleural & 1.20 (0.60, 3.25) & 99.999 \\
        Creatinine urine & 78.00 (24.00, 186.00) & 99.736 \\
        Diastolic blood pressure & 60.00 (44.50, 80.00) & 68.667 \\
        Eosinophils & 2.00 (1.00, 6.00) & 99.981 \\
        Fibrinogen & 220.00 (131.00, 434.00) & 98.713 \\
        Fraction inspired oxygen & 0.50 (0.40, 1.00) & 98.022 \\
        Fraction inspired oxygen Set & 0.50 (0.40, 1.00) & 97.423 \\
        Glascow coma scale total & 14.00 (3.00, 15.00) & 94.824 \\
        Glucose & 136.25 (95.00, 226.00) & 88.420 \\
        Heart Rate & 84.00 (63.00, 109.00) & 68.040 \\
        Height & 170.09 (154.97, 182.94) & 99.440 \\
        Hematocrit & 31.90 (24.10, 40.80) & 91.177 \\
        Hemoglobin & 10.80 (8.10, 13.90) & 91.924 \\
        Lactate & 2.00 (1.00, 4.88) & 96.639 \\
        Lactate dehydrogenase & 260.00 (156.00, 757.00) & 99.228 \\
        Lactate dehydrogenase pleural & 172.00 (65.20, 1485.00) & 99.991 \\
        Lactic acid & 2.10 (1.00, 5.10) & 98.029 \\
        Lymphocytes & 11.00 (3.80, 28.10) & 98.061 \\
        Lymphocytes ascites & 18.00 (2.00, 57.60) & 99.982 \\
        Lymphocytes atypical & 2.00 (1.00, 5.00) & 99.874 \\
        Lymphocytes atypical CSL & 1.50 (1.00, 3.00) & 99.998 \\
        Lymphocytes body fluid & 18.00 (2.00, 68.20) & 99.973 \\
        Lymphocytes percent & 13.00 (4.00, 39.00) & 99.993 \\
        Lymphocytes pleural & 30.00 (3.00, 79.40) & 99.991 \\
        Magnesium & 1.90 (1.50, 2.40) & 95.863 \\
        Mean blood pressure & 78.00 (61.00, 99.50) & 68.884 \\
        Mean corpuscular hemoglobin & 30.50 (27.50, 33.20) & 95.043 \\
        Mean corpuscular hemoglobin concentration & 34.10 (31.90, 35.80) & 95.040 \\
        Mean corpuscular volume & 89.00 (82.00, 97.00) & 95.043 \\
        Monocytes & 4.00 (1.70, 7.10) & 98.081 \\
        Monocytes CSL & 20.00 (4.00, 54.00) & 99.966 \\
        Neutrophils & 81.20 (60.00, 91.70) & 98.050 \\
        Oxygen saturation & 98.67 (94.00, 100.00) & 67.963 \\
        Partial pressure of carbon dioxide & 41.00 (32.00, 52.00) & 92.637 \\
        Partial pressure of oxygen & 195.00 (88.00, 387.00) & 97.213 \\
        Partial thromboplastin time & 30.60 (23.50, 55.50) & 94.662 \\
        Peak inspiratory pressure & 23.00 (14.00, 32.00) & 97.442 \\
        Phosphate & 3.50 (2.20, 5.30) & 96.903 \\
        Phosphorous & 3.40 (2.20, 5.20) & 97.871 \\
        Plateau Pressure & 18.00 (14.00, 25.00) & 98.226 \\
        Platelets & 203.00 (104.00, 352.00) & 93.685 \\
        Positive end-expiratory pressure & 5.00 (5.00, 10.00) & 99.095 \\
        Positive end-expiratory pressure Set & 5.00 (5.00, 8.00) & 96.835 \\
        Post Void Residual & 163.26 (65.40, 382.75) & 99.955 \\
        Potassium & 4.20 (3.40, 5.22) & 91.172 \\
        Potassium serum & 4.10 (3.40, 5.00) & 98.676 \\
        Prothrombin time INR & 1.30 (1.00, 2.00) & 94.650 \\
        Prothrombin time PT & 14.20 (12.20, 19.36) & 94.652 \\
        Pulmonary Artery Pressure mean & 26.00 (17.00, 40.00) & 98.976 \\
        Pulmonary Artery Pressure systolic & 33.00 (23.00, 47.00) & 93.292 \\
        Pulmonary Capillary Wedge Pressure & 16.00 (8.00, 27.00) & 99.902 \\
        Red blood cell count & 3.70 (2.75, 4.72) & 95.039 \\
        Red blood cell count CSF & 16.50 (1.00, 2950.00) & 99.966 \\
        Red blood cell count ascites & 1050.00 (43.20, 13450.00) & 99.982 \\
        Red blood cell count pleural & 3000.00 (182.60, 103866.40) & 99.991 \\
        Red blood cell count urine & 5.00 (1.00, 80.70) & 99.701 \\
        Respiratory rate & 17.00 (11.00, 25.00) & 67.945 \\
        Respiratory rate Set & 14.00 (10.00, 20.00) & 97.250 \\
        Sodium & 138.00 (133.00, 143.00) & 92.165 \\
        Systemic Vascular Resistance & 1037.04 (646.09, 1652.88) & 98.060 \\
        Systolic blood pressure & 117.00 (92.67, 150.00) & 68.656 \\
        Temperature & 36.56 (35.56, 37.61) & 88.330 \\
        Tidal Volume Observed & 564.00 (424.00, 800.00) & 96.496 \\
        Tidal Volume Set & 550.00 (450.00, 700.00) & 97.480 \\
        Tidal Volume Spontaneous & 460.00 (0.00, 714.00) & 99.135 \\
        Total Protein & 5.90 (4.83, 7.27) & 99.990 \\
        Total Protein Urine & 57.50 (13.00, 388.40) & 99.967 \\
        Troponin-I & 1.70 (0.30, 21.00) & 99.914 \\
        Troponin-T & 0.07 (0.01, 1.42) & 98.723 \\
        Venous PvO2 & 43.00 (29.00, 71.00) & 99.974 \\
        Weight & 77.80 (55.50, 108.08) & 97.047 \\
        White blood cell count & 11.10 (5.70, 20.00) & 94.052 \\
        White blood cell count urine & 4.00 (1.00, 40.90) & 99.706 \\
        pH & 7.37 (7.27, 7.46) & 92.402 \\
        pH urine & 5.50 (5.00, 7.00) & 98.473
        \label{tab:mimic_stats}
\end{longtable}
\end{small}

\clearpage
\begin{table}[h]
    \centering
    \caption{Percentage of positive samples for the extracted labels.}
    \begin{tabular}{lr}
         \toprule
         Label & Positive Class \\
         \midrule
         \texttt{vent} & 12.02\% \\
         \texttt{vaso} & 10.30\% \\
         \texttt{colloid\_bolus} & 0.95\% \\
         \texttt{crystalloid\_bolus} & 9.61\% \\
         \texttt{niv} & 37.74\% \\
         \bottomrule
    \end{tabular}
    \label{tab:mimic_labels}
\end{table}

\begin{table}[h]
    \centering
    \caption{Summary of the static variables for the extracted MIMIC-III patient cohort.}
    \begin{tabular}{lll}
         \toprule
         Static variable & & Patients \\
         \midrule
         \textbf{Sex} & Female & 14,978 (43\%) \\
          & Male & 19,494 (57\%) \\
          \midrule
          \textbf{Ethnicity} & Asian & 842 (2\%) \\
          & Hispanic & 1,137 (3\%) \\
          & Black & 2,667 (8\%) \\
          & Other & 5,183 (15\%) \\
          & White & 24,643 (71\%) \\
          \midrule
          \textbf{Age} & $<$30 & 1,832 (5\%) \\
          & 31-50 & 5,489 (16\%) \\
          & 51-70 & 12,942 (38\%) \\
          & $>$70 & 14,209 (41\%) \\
          \midrule
          \textbf{Insurance Type} & Self Pay & 477 (1\%) \\
          & Government & 1,050 (3\%) \\
          & Medicaid & 2,782 (8\%) \\
          & Private & 11,846 (34\%) \\
          & Medicare & 18,317 (53\%) \\
          \bottomrule
    \end{tabular}
    \label{tab:mimic_static}
\end{table}

\section{Samples of synthetically generated data}
\label{app:synth_examples}

\begin{sidewaysfigure}
    \centering
    \includegraphics[width=\textwidth]{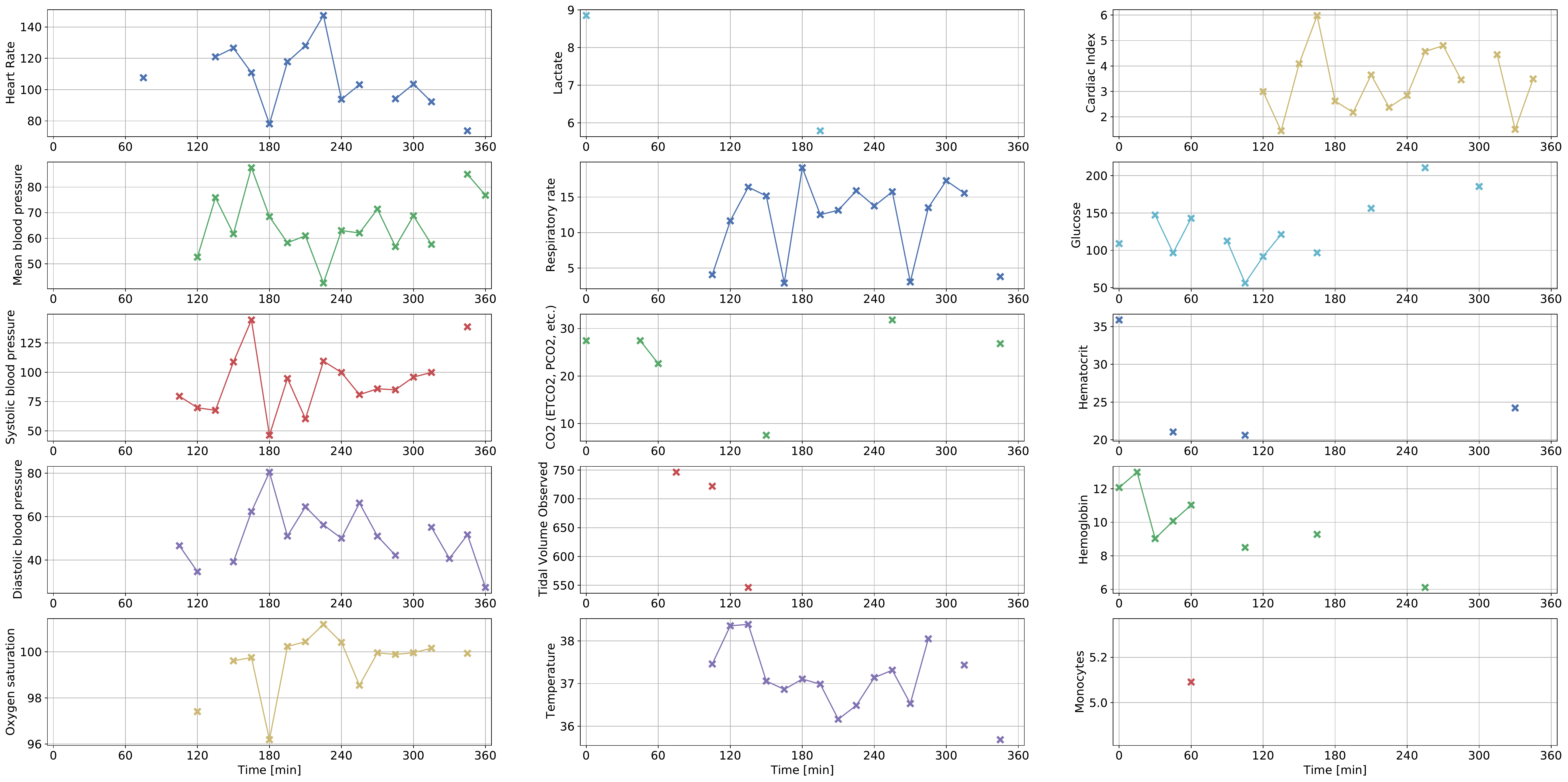}
    \caption{Feature time series of a synthetically generated patient.}
    \label{fig:synth_sample_1}
\end{sidewaysfigure}

\begin{sidewaysfigure}
    \centering
    \includegraphics[width=\textwidth]{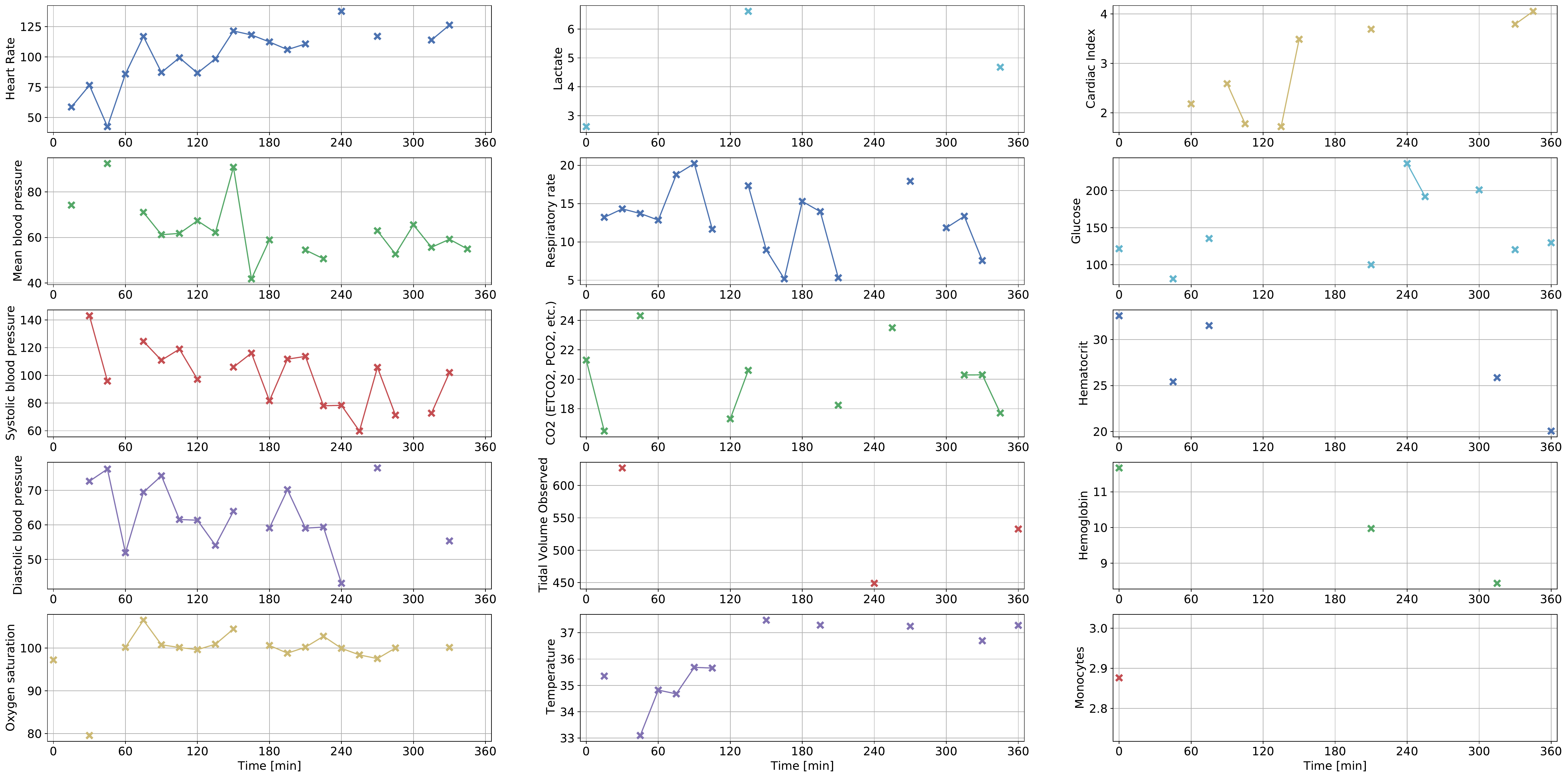}
    \caption{Feature time series of a synthetically generated patient.}
    \label{fig:synth_sample_2}
\end{sidewaysfigure}

\begin{sidewaysfigure}
    \centering
    \includegraphics[width=\textwidth]{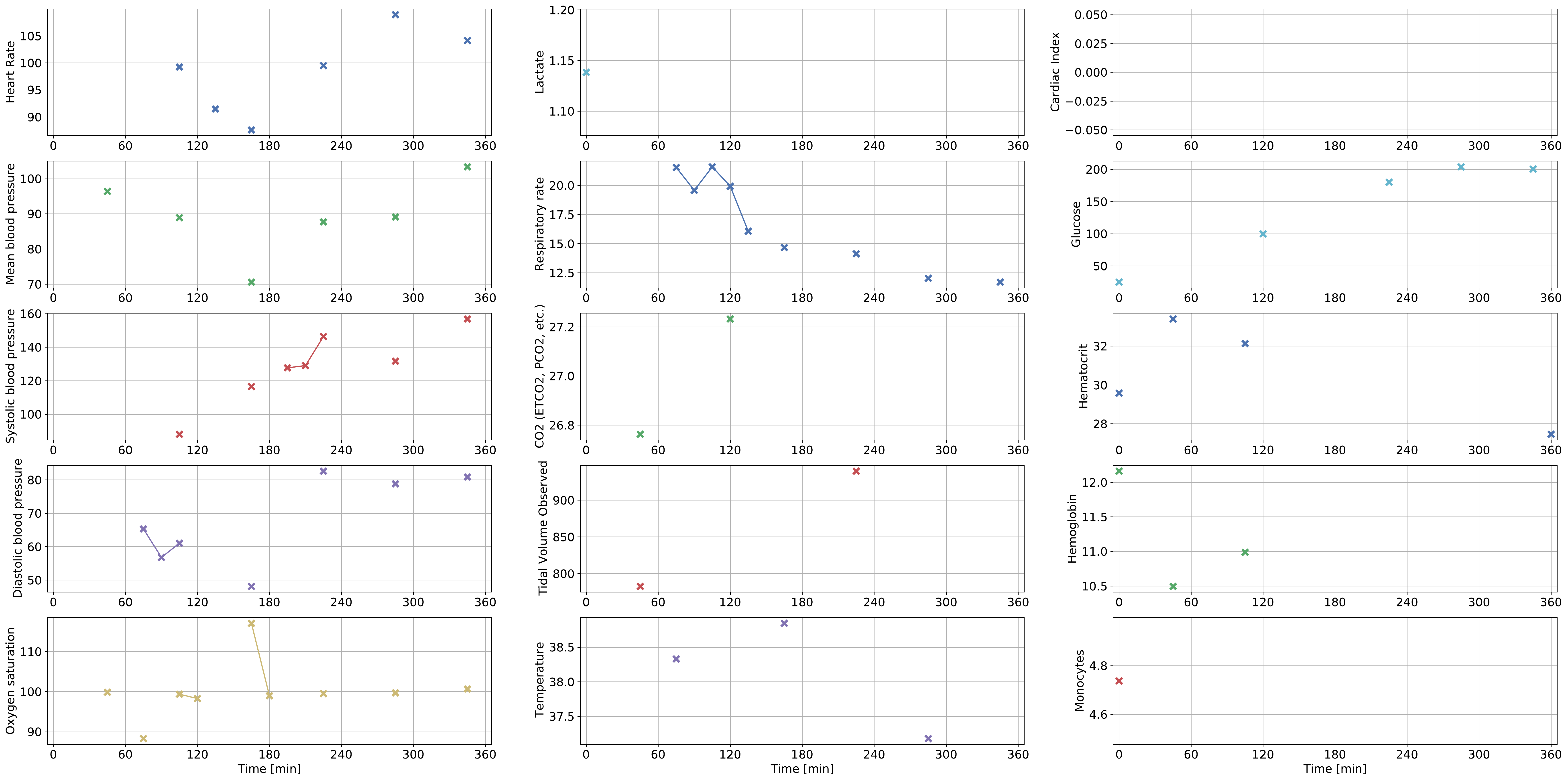}
    \caption{Feature time series of a synthetically generated patient.}
    \label{fig:synth_sample_3}
\end{sidewaysfigure}

\begin{sidewaysfigure}
    \centering
    \includegraphics[width=\textwidth]{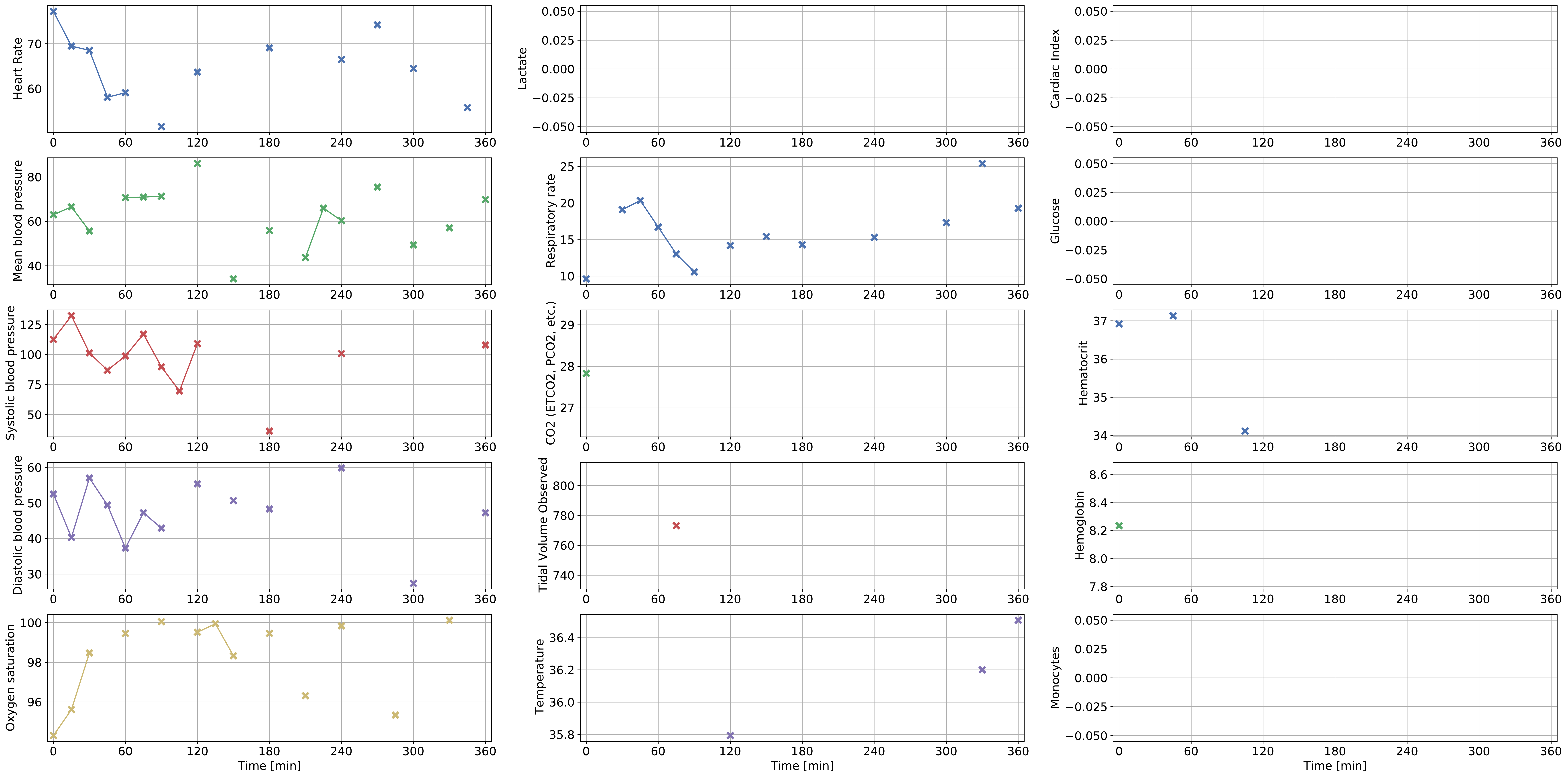}
    \caption{Feature time series of a synthetically generated patient.}
    \label{fig:synth_sample_4}
\end{sidewaysfigure}

\begin{sidewaysfigure}
    \centering
    \includegraphics[width=\textwidth]{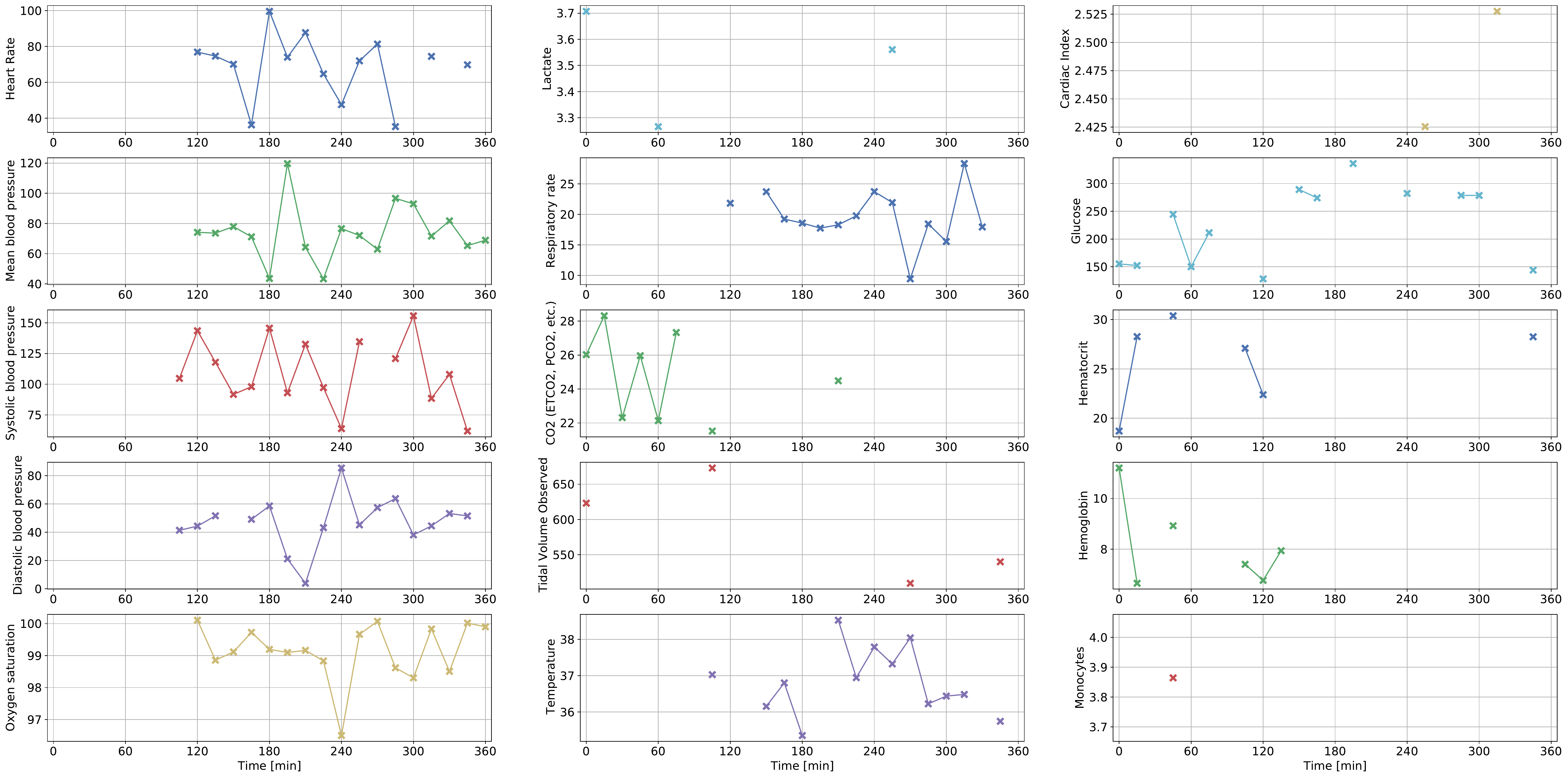}
    \caption{Feature time series of a synthetically generated patient.-}
    \label{fig:synth_sample_5}
\end{sidewaysfigure}

For the additional synthetic data visualization we preform a preselection of the generated patients based on lower than average missingness, in order to visualize samples where dense measurements over time are present and the dynamics are more evident .
After this preselection, we randomly select three patients to visualize amongst them in \Cref{fig:synth_sample_1,,fig:synth_sample_2,,fig:synth_sample_5}. We also show two examples of randomly selected patients without this preselection in \Cref{fig:synth_sample_3,,fig:synth_sample_4}.

\end{document}